%% file: main.tex
\documentclass{article} 
\usepackage{iclr2026_conference,times}
\usepackage[T1]{fontenc}

\input{math_commands.tex}

\usepackage{hyperref}
\usepackage{url}
\usepackage{amsthm}
\usepackage{cleveref}
\usepackage{graphicx}
\usepackage{listings,xcolor}
\usepackage{why3lang}
\usepackage{subcaption}
\usepackage{wrapfig, booktabs}
\usepackage{multirow}
\usepackage{xspace}
\usepackage[most,skins,theorems]{tcolorbox}
\usepackage{pifont}
\usepackage{mathtools}
\hypersetup{
  colorlinks=true,          
  pdfborder={0 0 0},        
  linkcolor=purple!50!black,
  citecolor=purple!50!black,
  urlcolor=purple!50!black, 
  filecolor=purple!50!black 
}

\DeclareMathOperator{\xlen}{length}
\newcommand{\xif}{\mathrm{if}}
\newcommand{\xthen}{\mathrm{then}}
\newcommand{\xelse}{\mathrm{else}}

\definecolor{codebg}{HTML}{F7F7F9}
\definecolor{kw}{HTML}{005CC5}      
\definecolor{ty}{HTML}{6F42C1}      
\definecolor{cm}{HTML}{6A737D}      
\definecolor{st}{HTML}{0F766E}      
\definecolor{nm}{HTML}{B31D28}      

\newcommand{\xmark}{{\color{red!65!black}\ding{55}}} 
\newcommand{\cmark}{{\color{green!40!black}\ding{51}}}

\lstdefinestyle{whyml-light}{
  language=why3,
  basicstyle=\ttfamily\small,
  keywordstyle=\color{kw}\bfseries,
  commentstyle=\itshape\color{cm},
  stringstyle=\color{st},
  numberstyle=\scriptsize\color{cm},
  numbers=left, numbersep=8pt,
  showstringspaces=false, tabsize=2,
  upquote=true
}
\crefname{part}{\S}{\S\S}
\crefname{chapter}{\S}{\S\S}
\crefname{section}{\S}{\S\S}
\crefname{subsection}{\S}{\S\S}

\crefname{table}{Tab.}{Tabs.}
\crefname{figure}{Fig.}{Figs.}

\theoremstyle{definition}
\newtheorem{definition}{Definition}

\newcommand{\joshcommand}[1]{}

\title{Neural Theorem Proving for Verification\\Conditions: A Real-World Benchmark}

\author{
Qiyuan Xu\textsuperscript{1,3}, Xiaokun Luan\textsuperscript{2}, Renxi Wang\textsuperscript{3}, Joshua Ong Jun Leang\textsuperscript{4,6}, Peixin Wang\textsuperscript{5},\\
\ \textbf{Haonan Li\textsuperscript{3}\thanks{Corresponding Author}, Wenda Li\textsuperscript{6}, Conrad Watt\textsuperscript{1}}\\[.4em]
\textsuperscript{1} Nanyang Technological University\qquad
\textsuperscript{2} Peking University\qquad
\textsuperscript{3} MBZUAI \\
\textsuperscript{4} Imperial College London\quad
\textsuperscript{5} East China Normal University\quad
\textsuperscript{6} University of Edinburgh
}

\tcbset{
  aibox/.style={
    width=\linewidth,
    top=7pt,
    bottom=2pt,
    colback=blue!6!white,
    colframe=black,
    colbacktitle=black,
    enhanced,
    breakable, 
    center,
    attach boxed title to top left={yshift=-0.1in,xshift=0.15in},
    boxed title style={boxrule=0pt,colframe=white,},
    before upper=\setlength{\parskip}{2pt}\setlength{\parindent}{0pt}, 
  }
}
\newtcolorbox{AIbox}[2][]{aibox,title=#2,#1}

\iclrfinalcopy 
\begin{document}

\maketitle

\begin{abstract}

Theorem proving is fundamental to program verification, where the automated proof of Verification Conditions (VCs) remains a primary bottleneck. Real-world program verification frequently encounters hard VCs that existing Automated Theorem Provers (ATPs) cannot prove, leading to a critical need for extensive manual proofs that burden practical application. While Neural Theorem Proving (NTP) has achieved significant success in mathematical competitions, demonstrating the potential of machine learning approaches to formal reasoning, its application to program verification—particularly VC proving—remains largely unexplored. 
Despite existing work on annotation synthesis and verification-related theorem proving, no benchmark has specifically targeted this fundamental bottleneck: automated VC proving.
This work introduces \textbf{Neural Theorem Proving for Verification Conditions (NTP4VC)}, presenting the first real-world multi-language benchmark for this task. From real-world projects such as Linux and Contiki-OS kernel, our benchmark leverages industrial pipelines (Why3 and Frama-C) to generate semantically equivalent test cases across formal languages of Isabelle, Lean, and Rocq. We evaluate large language models (LLMs), both general-purpose and those fine-tuned for theorem proving, on NTP4VC. Results indicate that although LLMs show promise in VC proving, significant challenges remain for program verification, highlighting a large gap and opportunity for future research.

\end{abstract}

\input{iclr2026/sections/intro_2}

\input{iclr2026/sections/background}
\input{iclr2026/sections/method}

\input{iclr2026/sections/benchmark}
\input{iclr2026/sections/experiment}
\input{iclr2026/sections/related_works}

\section{Conclusion}

This work introduces Neural Theorem Proving for Verification Conditions (NTP4VC), presenting the first real-world multi-language benchmark for automated VC proving --- a critical bottleneck in program verification.
Alongside this benchmark, this work develops a reliable extraction method using expert-written translation rules and industrial verification pipelines (Why3 and Frama-C) to extract VC corpora from real-world verification projects and generate semantically equivalent VCs across Isabelle, Lean, and Rocq.
Our evaluation of 600 carefully selected VCs from industrial projects reveals the substantial difficulty of this task: the strongest neural theorem provers achieve only 2.08\% pass@1.
Our error analysis reveals that the lengthy, deeply nested structure of VCs presents fundamentally different challenges to NTP models compared to mathematics competition problems.
The benchmark and the corpora extraction method establish a foundation for advancing neural approaches to program verification, with the potential to achieve significant breakthroughs in automated program verification.




\section*{Reproducibility Statement}
We have made a comprehensive artifact to ensure the reproducibility of our results and to encourage future research. The artifact contains the complete NTP4VC benchmark, the source code for our VC extraction tool, and all scripts required to replicate our experiments.
It is available at:
\begin{center}
  \url{https://github.com/xqyww123/NTP4VC}
\end{center}

\bibliography{iclr2026_conference}
\bibliographystyle{iclr2026_conference}

\appendix

\input{iclr2026/sections/appendix}

\end{document}

%% file: math_commands.tex
\usepackage{amsmath,amsfonts,bm}

\def\eqref#1{equation~\ref{#1}}

\def\1{\bm{1}}

\DeclareMathAlphabet{\mathsfit}{\encodingdefault}{\sfdefault}{m}{sl}
\SetMathAlphabet{\mathsfit}{bold}{\encodingdefault}{\sfdefault}{bx}{n}



%% file: iclr2026/sections/intro_2.tex
\section{Introduction}\label{sec:intro}

Program verification has been fundamental to software reliability for over half a century~\citep{HoareTriple}. While numerous industrial program verifiers have been developed and deployed in history~\citep{cousot2005astree}, the adoption of program verification remains limited to safety-critical domains~\citep{rushby2009software,woodcock2009formal}.
A primary reason is the heavy manual effort required in the theorem proving of \emph{Verification Conditions (VCs)}~\citep{Boogie}: the logical propositions that encode program correctness.

VC plays a central role in the conventional workflow of program verification~\citep{VCC,Dafny} as shown in \cref{fig:verification}: the Verification Condition Generator (VCG) component aims to generate VCs and the prover aims to prove them.
Conventionally, VC proving is carried out by Automated Theorem Provers (ATPs).
However, ATPs excel only at specific domains of problems, and require human intervention (e.g., manual proofs and annotations) when automatic proof attempts fail or time out. 
Taking the widely-used industry tool Frama-C~\citep{FramaC} as an example, existing ATPs' insufficient capability necessitates $\sim$600 lines of annotations for a linked list library, nearly matching the original C code length.
Consequently, due to the central role of VC proving and the inadequacy of current automated approaches, VC proving has become \emph{a key bottleneck} in automated program verification.



Large language models (LLMs) have opened the door to Neural Theorem Proving (NTP)~\citep{minervini2018towards}, where models generate formal proofs to conduct theorem proving. While existing NTP research has focused primarily on mathematical domains, proving competition problems~\citep{MiniF2F,PutnamBench} and formalizing mathematics~\citep{APE}, theorem proving extends naturally to VC proving~\citep{HARRISON2014135}.


\input{iclr2026/figures/VCG}


This motivates our central question: \emph{can NTP automate VC proving?} To answer this, we introduce \textbf{Neural Theorem Proving for Verification Conditions (NTP4VC) --- a task that applies machine-learning-based proof generation to conduct the theorem proving of VCs.}


To evaluate this task, we construct the first benchmark for NTP4VC, whose major features are compared with prior works in \cref{tab:comparison}.
A challenge of this construction is that Lean~\citep{Lean4}, a mainstream language in the NTP community, has relatively fewer mature program verification frameworks built on top of it and large-scale industrial verification projects using it.
Despite our best efforts, we find no sufficient native VCs available in Lean for a NTP4VC benchmark.

We overcome this issue by translating the VCs generated from other industrial verification pipelines (Why3~\citep{Why3} and Frama-C~\citep{FramaC}) into Lean.
This approach also allows us to translate VCs to Isabelle ~\citep{Isabelle}, Rocq~\citep{CoquandHuet1988CoC} (which are already implemented), and potentially other target languages, forming the first multi-language benchmark in NTP-based program verification.
More crucially, this approach further allows extracting VCs from existing verification projects for industrial software, such as the Linux kernel's scheduler{~\citep{ShouldWeBalance}, library functions~\cite{verker},} and Contiki OS's memory allocator~\citep{ContikiMem} and linked-list library~\citep{ContikiList}.

Unlike LLM-based translation approaches that suffer from LLMs' unreliability, our translation pipeline is based on $\sim$800 expert-written translation rules for each of the three target languages (so ${\sim}3 \times 800$ in total). These rules are explicitly chosen to ensure semantic preservation from the origins to the translations, thereby better ensuring the quality of the benchmark cases compared to LLM-based translation approaches.



\joshcommand{Draft: We conduct comprehensive evaluation of both specialised theorem-proving models and general-purpose large language models on NTP4VC. Specialised theorem-proving models, fine-tuned for formal reasoning tasks, achieve modest success with the top-performing model reaching 30\% pass@1. However, general-purpose LLMs perform markedly worse, with GPT-4o-mini obtaining only 10\% pass@1.  These baseline results demonstrate that while neural approaches show promise for VC proving, current capabilities remain insufficient for practical deployment, highlighting the need for targeted advances in neural theorem-proving methodologies. -> I would include anlaysis too here. I'll refine here later on
}
We further evaluate several existing provers and LLMs on NTP4VC. For language-specific fine-tuned provers, the best model achieves only 2.08\% pass@1, while general-purpose LLMs achieve lower performance, with GPT-o4-mini-high achieving 1.19\% pass@1. These results highlight the substantial difficulty of VC proving and the need for progress in NTP and LLM reasoning.

To summarize, our contribution includes:
\begin{enumerate}
\item We define the task of NTP4VC (\cref{sec:intro}), which aims to attack the automated proving of VC, a key bottleneck in program verification.
\item We propose a \emph{reliably automatic} method for extracting corpora from real-world verification projects (\cref{sec:method}). The implementation is open-sourced.
\item We present the first real-world, multi-language benchmark for NTP4VC, with open-sourced implementation and extensive evaluation of existing provers and LLMs (\cref{sec:evaluation}).
\end{enumerate}

\input{iclr2026/tables/comparison}
\input{iclr2026/figures/pipeline}

%% file: iclr2026/figures/VCG.tex
\begin{figure}[t]
    \centering
    \includegraphics[width=\linewidth]{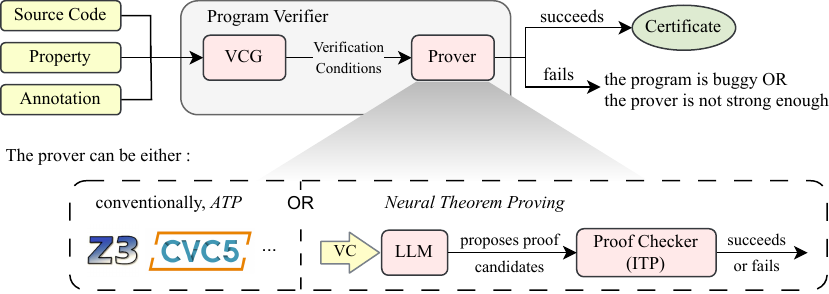}
    \caption{The conventional and NTP-based workflow of program verification.}
    \label{fig:verification}
\end{figure}

%% file: iclr2026/tables/comparison.tex
\newcommand{\xhead}[2]{\multirow{2}{*}{\parbox[c][0.7cm][c]{#1}{\bf #2}} }
\begin{table}[t]
    \vspace*{-1em}
    \centering
    \caption{Comparison between our benchmark and previous ITP-based benchmarks for program verification. \textbf{VC}: the proportion of VC test cases. \textbf{Industrial pipeline}: whether the work uses industrial program verification pipelines. \textbf{Language}: the proof language supported by the benchmark.}
\label{tab:comparison}
\small%
\begin{tabular}{llrcccc}
\toprule
\xhead{1.5cm}{Benchmarks} & \xhead{1cm}{Focus} & \xhead{.6cm}{\centering VC} & \multirow{2}{*}{\!\!\!\parbox[c][.8cm][c]{1.5cm}{\bf\centering Indstrial\\[.2em]Pipeline}}\!\!\! & \multicolumn{3}{c}{\bf Language} \\
\cmidrule(lr){5-7}
 & & & & \bf Lean & \bf\!\!Isabelle\!\! & \bf Rocq \\
\midrule
\citet{FVEL} & \multirow{2}{*}{\parbox{2.2cm}{\small verificatoin-related lemmas}} & $< 17\%$ & \cmark & \xmark & \cmark & \xmark \\
\citet{Rango} &  & $< 20\%$ & \cmark & \xmark & \xmark & \cmark \\
\cmidrule(lr){1-2}
\citet{CLEVER} & \multirow{2}{*}{\parbox[c][1cm][c]{2.2cm}{\small programming puzzles in Lean}} & 0\% & \xmark & \cmark & \xmark & \xmark \\
\citet{DoughertyMehta2025} &  & 0\% & \xmark & \cmark & \xmark & \xmark \\
\citet{miniCodeProps} & & 0\% & \xmark & \cmark & \xmark & \xmark
\\
\cmidrule(lr){1-2}
Ours & \parbox{2.6cm}{\small VCs from puzzles \& industrial projects} & 100\% & \cmark & \cmark & \cmark & \cmark \\
\bottomrule
    \end{tabular}
\normalsize
\end{table}

%% file: iclr2026/figures/pipeline.tex
\begin{figure}[t]
    \centering
    \includegraphics[width=\linewidth]{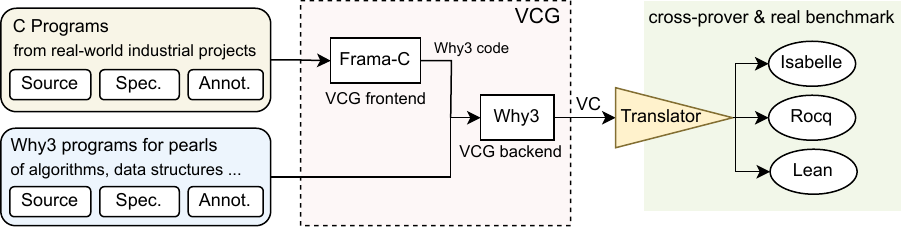}
    \caption{Our pipeline for extracting benchmark cases.
}
    \label{fig:pipeline}
\end{figure}

%% file: iclr2026/sections/background.tex
\section{Background}\label{sec:background}

Theorem proving falls broadly into two categories: \textbf{Automated Theorem Proving (ATP)} and \textbf{Interactive Theorem Proving (ITP)}.
ATP achieves full automation within specific domains of proof problems. These domains are limited, and VCs in real-world verification projects often exceed these domains, leading to proof failures and inevitable human intervention (e.g., manual proofs and annotations) in order to complete the proofs.
By contrast, ITP provide highly expressive languages that enable users to construct proofs across broad domains, capable of handling almost all program verification problems.
Mainstream ITP languages include Isabelle, Rocq, and Lean.


\textbf{Program verification} aims to verify that a program satisfies a given property.
Ideally, a strong enough verifier should be able to complete the verification solely given the source code and the property.
In practice, however, due to limitations in both VCG and VC provers, users often have to provide manual proofs and annotations to guide the verifier in completing the verification.
The manual effort for these proofs and annotations constitutes a huge cost burden in program verification.

\textbf{Why3} and \textbf{Frama-C} are famous program verifiers widely used in the industry.
Why3 provides 1) a language for both programming, annotation, and specifying functional correctness, 2) a VCG, and 3) powerful ATPs.
A limitation is that Why3 can only verify programs written in its abstract specification language. In order to verify programs written in industrial languages, Why3 is widely used as the verification backend of well-known toolchains which translate their input language to the Why3 specification language --- including Frama-C, Cameleer~\citep{Cameleer}, Creusot~\citep{Creusot}, and EasyCrypt~\citep{EasyCrypt}.
Frama-C is an industrial verifier for the C language.
It provides a frontend to process C source code and then calls Why3 to complete the verification.
The input of Frama-C is C source code with properties and annotations provided as comments, and the output is Why3 code that Why3 can continue to verify.
Finally, Frama-C is widely used, having verified enormous industrial programs, such as air traffic management algorithm~\citep{Dutle2021}, embedded operating system ~\citep{ContikiMem,ContikiList}, cryptographic modules~\citep{AESCCM}, Linux kernel scheduler~\citep{ShouldWeBalance}, and JavaCard virtual machine~\citep{JavaCard}.

%% file: iclr2026/sections/method.tex
\section{A Reliable and Automatic Method for Corpora Generation}\label{sec:method}


This section presents the method we use to extract real-world VCs that consitute our benchmark.
The key idea is to reuse existing industrial VCGs to extract VCs from existing verification projects, and translate these VCs into the language of the target ITPs (\cref{sec:translation}).
Since the projects have all passed the verifiers’ checks, the VCs are guaranteed to be provable. However, this also makes them too easy to serve as valuable benchmark cases, as they may already be within reach of existing ATPs. To produce challenging benchmark cases, we introduce a novel complication process (\cref{sec:complication}) to make VCs harder while keeping them provable. The complete process is illustrated in \cref{fig:translation}.



\subsection{VC Extraction \& Rule-based Translation}\label{sec:translation}

Various VC languages are used in industry, such as Why3, TPTP~\citep{TPTP}, and SMT Lib~\citep{SMT-Lib}.
We adopt Why3 because its logic system is Simple Typed Theory~\citep{Church_1940}, a relatively high-level system that is close to and entailed by the logic of mainstream ITPs like Lean, Isabelle, and Rocq, ensuring the feasibility of the translation.

\input{iclr2026/figures/translation_process}

The translation process begins with a given Why3 source code. It first runs Why3 VCG to generate VCs and calls our customized Why3 printer to dump the VCs into an XML representation of their Abstract Syntax Trees (ASTs).
These ASTs are processed by a Python translation framework also written by us and finally mapped into the target ITPs' languages. The details are provided in \cref{appendix:extraction}.

While the above process enables the basic translation from Why3 to target ITP languages, our work goes beyond this to strive for idiomatic translations that closely approximate native expressions on the target ITP platforms.
For this, our translation process incorporates enhancements from two aspects:
First, at the syntactic level, we use printing rules to map specific term structures to their corresponding pretty syntax defined in the ITP, including prefix, infix annotations, and ad-hoc syntax sugars like if-then-else, match-case, and list[index].
Second, we build a rewriting system to rewrite specific combinations of terms into more idiomatic expressions.
Examples include rewriting integer operations into natural number operations that are more common in ITP.

The implementation of the pipeline is made of more than 2400 mapping \& rewriting rules written by human experts in ITP, in total for Isabelle, Lean, and Rocq.
The correctness of the rules is supported by syntax checking over the translation results on one hand, and cross-validation by other experts (our first, second, and last authors) on the other hand.
These expert-written rules form the foundation of the translations' correctness and quality.
Once this foundation is built, the entire translation process is automatic, constituting a \emph{reliably automatic} method for extracting VC corpora.

\subsection{Complication Process: Extracting Challenging VCs}\label{sec:complication}

As mentioned at the beginning of this section, the VCs extracted from real-world projects are already provable by existing ATPs, thus providing insufficient challenge for benchmark evaluation.
However, these VCs are provable by the ATPs as human developers have already written sufficient annotations to make them easy for ATPs to prove, rather than from inherent ATP strength. A direct idea is to erase these auxiliary annotations and restore the verification tasks to what they should ideally be in fully automated program verification.

Specifically, three sorts of annotations are dedicated to VC simplification: (1)
\texttt{assert} annotation, which introduces a subgoal to ask the prover to first prove this subgoal and then use the proven subgoal as a lemma in the subsequent proofs;
(2) \texttt{lemma} annotation, which explicitly introduces a global lemma so that the prover can later reference it to prove subsequent propositions;
(3) annotation of lemma application, which explicitly instantiates (the free variables in) a lemma and advises the prover to use it.
All these annotations can be safely erased without affecting the VC's provability (by a strong enough prover)~\citep{Why3Manual,FramaCManual}.
In addition, they exhibit clear syntactic patterns enabling us to identify and erase them.
Indeed, the exact job of our complication process is erasing the annotations.
The results show this process effectively reduces the pass rate of Why3's strongest ATP from $\sim$99\% to $\sim$62\% on Why3's bundled examples.

%% file: iclr2026/figures/translation_process.tex
\begin{figure}[t]
    \centering
    \includegraphics[width=\linewidth]{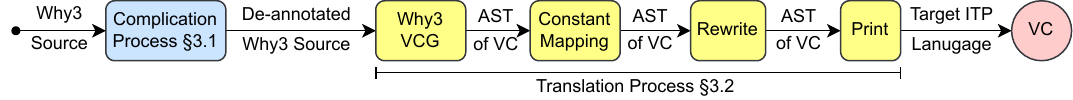}
    \caption{The generation process of the benchmark cases and potentially training corpora.}
    \label{fig:translation}
\end{figure}

%% file: iclr2026/sections/benchmark.tex
\section{NTP4VC Benchmark}\label{sec:benchmark}

The method discussed in \cref{sec:method} enables effective extraction of real-world VCs from existing verification projects.
By applying the method, we extract $>$7.5k VCs from various sources.
From there, we carefully select 600 VCs to constitute a benchmark, with consideration for breadth, diversity, and the balance of difficulty levels as described below.

\begin{wraptable}{r}{5.5cm}
\vspace{-\intextsep}
\centering
\small
\caption{Categories of the benchmark}\label{tab:categories}
\vspace*{-1em}
\begin{tabular}{lcc}\\\toprule  
Category & \!\!Number & ATP pass \\\midrule
\multicolumn{3}{l}{\it pearls of programs}\\[.2em]
Algorithm       & 55 & 20.00\%\\[.1em]
Data Structure\!\!\! & 73 & 19.18\%\\[.1em]
Calculation     & 66 & 19.70\%\\[.1em]
Engineering     & 54 & 20.37\%\\[.1em]
Competition     & 52 & 19.23\%\\[.2em]
\multicolumn{3}{l}{\it real C verification}\\[.2em]
Function        &  81 & 23.46\%\\[.1em]
Loop            &  81 & 24.69\%\\[.1em]
Memory          &  74 & 24.32\%\\[.1em]
Invalid Arg.    &  64 & 25.00\%\\\midrule
Total           & 600 & 22.00\% \\
\bottomrule
\end{tabular}
\end{wraptable} 

\noindent\textbf{Breadth, Diversity, and the Difficulty Level.}
Real-world industrial projects certainly possess high value in a verification benchmark like ours, while at the same time, challenging algorithms and data structures are equally valuable verification targets due to their complexity.
An issue is that a conflict exists between them: challenging algorithmic content is sparse in industrial project source code.
If a benchmark focused solely on VCs from industrial projects, it would underrepresent algorithms and data structures.
In order to balance the breadth of the verification scenarios involved, we divide our benchmark into two equal parts (50\% vs 50\%).
(1) \textbf{Pearls of Programs} consists of minimal working programs that capture verification pain points, including algorithms, data structures, and well-known ``hard nuts to crack'', such as Binomial Heap, VerifyThis'24 competition, and Hillel challenge~\citep{Hillel}. These programs are written in Why3's abstract specification language.
(2) \textbf{Real C Verification}: VCs from industrial C programs used in real-world projects, such as the memory allocator~\citep{ContikiMem} and the linked-list library~\citep{ContikiList} from the Contiki Operating System.

Each category is further divided into sub-categories (\cref{tab:categories}), with roughly balanced numbers of cases in each sub-category to maintain diversity.
The pearl of programs consists of
1) Well-known algorithms such as sorting, string operations, searching, shortest path, and graph;
2) Data structures, including (balanced) trees, heaps, hash, and arrays;
3) Numerical and other calculations, such as arbitrary precision arithmetics, square root, exponentiation by squaring, and bitwise operations;
4) Engineering optimization tricks (e.g., in-place reversal of linked lists and N-queens by bitvector) and common engineering tasks (e.g., string padding, list element removal, space-insensitive comparison between strings, and the challenges by~\cite{Hillel});
5) Cases from well-known verification competitions, e.g., 
VerifyThis~\citep{VerifyThis} and VSCOMP~\citep{VSCOMP}.

While the pearl of programs is organized by source programs' functionality, cases in the real C verification are categorized by the properties that VCs validate:
1) {Function} category verifies that programs' logical behavior meets the desired functionalities from a big-picture view, assuming the absence of runtime errors;
2) {Loop} category verifies loop termination, and loop invariants are established and maintained;
3) {Memory} category rules out the runtime error of invalid memory access;
4) {Invalid Arg.} checks that arguments and operands are valid.
For example, the operands of multiplication do not cause arithmetic overflow, and the dividend is not zero.

Beyond breadth, we design the benchmark to balance difficulty across categories.
We measure difficulty using the pass rate of Why3’s strongest predefined ATP tactic, \texttt{Auto\_Level\_3} (AL3)~\citep{AL3}.
AL3 is a hybrid tactic that combines sophisticated heuristics and five industrial cutting-edge ATPs, Z3~\citep{Z3}, CVC4~\citep{barrett2011cvc4}, SPASS~\citep{SPASS}, Alt-Ergo~\citep{AltErgo}, and E-prover~\citep{Eprover}, such that a goal is proved once any of the ATPs proves the goal.
The pass rate of AL3 then indicates the state-of-the-art of Why3 ATP over the benchmark cases, denoted as \emph{ATP pass@n} in \cref{tab:categories}.
A VC is deemed \emph{hard} if AL3 fails to prove it, making its solution an open problem. Our design goal is to set each category’s composition to target an AL3 pass rate between 20\%$-$25\%, a level that ensures sufficient open problems for advancing NTP while still allowing effective evaluation of existing approaches.



\noindent\textbf{Diversity of VC Expressions}
\input{iclr2026/tables/VC-expr}
While the previous subsection measures the diversity of the source and the purpose of the VCs, this subsection discusses the arithmetic and data structure operations involved in these VCs.
We follow the taxonomic methodology conventionally used in the ATP field~\citep{SMT-Lib, smtlib_logics}, and categorize the operations according to the notions and the data types involved in their related reasoning.
As listed in \cref{tab:VC_stat}, the categories include integer arithmetics, non-linear arithmetic%
, and various common data structures. Some cases may define their custom datatypes beyond those provided in the standard libraries. This is captured by the \emph{Custom datatype} category. Further details about this classification are given in \cref{appendix:VC_stat_detail}.

For each of the categories, we count the benchmark cases that involve at least one such operation, and report the average, $25^\mathrm{th}$, and $75^\mathrm{th}$ percentile of: \emph{\# of operations}, the total number of occurrences of these operations; \emph{\# of unique oprs}, the number of distinct operation types in each case in each case; \emph{size}, the number of atomic terms; \emph{depth}, the height of the abstract syntax tree of the VCs;
\emph{\# of $\forall\exists$}, the number of quantifiers occurring in each case.
As presented in \cref{tab:VC_stat}, the result shows our benchmark cases exhibit a wide distribution across different data structures and arithmetic operations, and also span VCs of varying scales within each category.

\noindent\textbf{Sources of the Benchmark Cases, and Their Licenses.}
All the VCs in the benchmark are drawn from open-sourced verification projects.
The pearls of programs come from the Gallery of Verified Programs~\cite{toccata_gallery}, released under the LGPL v2.1 license alongside Why3’s source code.
For real C verification, VCs are collected from multiple projects, as summarized in \cref{tab:source}.
\input{iclr2026/tables/source}

\noindent\textbf{VC Selection Process.}
The 600 benchmark cases are selected from over 7.5k VCs. This subsection elaborates on the selection process.
The process consists of three rounds: The first round determines the domain from which the benchmark cases will be selected; in the second round, one expert performs an initial screening to identify ${\sim}1.5$k candidate cases; three experts then collaboratively evaluate each candidate in the final round to finalize the benchmark set of 600 cases.


Recall that cases in Pearls of Programs are sourced from the \cite{toccata_gallery}'s collection of 224 individual projects. In the first round, we select 100 projects from which all Pearl benchmark cases will be drawn, leaving the remaining 124 projects for potential use as training data.
To collect as many hard VCs as possible, we prioritize selecting projects rich in hard VCs. To measure if a VC is hard, we run Why3's AL3, and it is hard if and only if AL3 fails to solve it in 10 minutes on a 12-core workstation.
For Real C Verification, we do not maintain such project-level separation, and the 8 projects are all used for benchmark cases.

In the second round, we first select all the hard VCs from the domain, totaling ${\sim}1.2$k cases.
Since we aim for the benchmark to have a $20\text{–}25$\% pass rate on the ATP baseline, we correspondingly select ${\sim}350$ cases from the easy VCs to balance the candidate set at this stage.
When selecting each easy VC, we check whether it is trivially provable (e.g., $\mathrm{true} \land \mathrm{true}$). To do this, we examine the VC's logical expression, the source program, related annotations, and the specifications to check that the property verified by the VC is meaningful and commonly encountered in program verification tasks.

In the final round, we apply the same evaluation method above to assess each case and refine the candidate set while additionally considering balanced coverage across the categories shown in \cref{tab:categories}.
We also consider broad project coverage by selecting cases from different projects proportionally.

\noindent\textbf{Format of the Benchmark Cases.}
Each benchmark case is a single VC (a single proof goal) placed individually in a theory file, and each such file contains exactly one VC. Every VC originates from a verification project and thus may contain project-specific concepts (e.g., the data type of binary tree), resulting in VCs with library dependencies. Consequently, this requires benchmark participants to be able to learn new concepts on-the-fly from the verification projects' dependency libraries.

\noindent\textbf{Dataset Contamination.}
Our benchmark is generally free of data contamination concerns, despite all the source programs, properties, and annotations are public.
This is because: (1) The transformation from program and property source code to VCs is complex. Even if LLMs were trained on the original source code, they cannot trivially generate VC-level concepts.
In typical program verification workflows, VCs are generated only transiently and are not persistently stored or published unless done deliberately.
(2) The VCs we use are derived from Why3 source code after a complication process, making most of them unprovable by existing ATPs, proofs for these VCs have never existed.
(3) Even if we assume the proof details of the VCs from the original Why3 source code can leak information about the proofs of the complicated Why3 code, no leakage of the proof details is discovered despite our best efforts.
This is expected, since Why3 never stores detailed proofs, but only 
records the ATP tools used, replaying them when proofs are needed.
In fact, many ATPs do not support dumping detailed proofs at all.
In summary, the risk of meaningful data contamination in our benchmark is extremely low.

\noindent\textbf{Caveat: Provability of Benchmark Cases.}
All benchmark cases are drawn (after the complication process) from verification conditions in the projects that are formally verified.
Methodologically, this should guarantee the provability of all benchmark cases.
However, in practice, all VCs are processed through Frama-C, Why3, and our translation pipelines. Implementation bugs in any of these components may potentially render some benchmark cases unprovable. During our selection process, we checked the provability of each case to the extent possible. However, due to the inherent complexity of theorem proving, we cannot guarantee the actual provability of every case. Therefore, we must include this caveat in the paper regarding the provability of benchmark cases.

%% file: iclr2026/tables/VC-expr.tex
\begin{table}[t]
\centering%
\small
\caption{Sources of cases in real C verification. LoC = Lines of C Code (comments are excluded).}\label{tab:source}
\vspace*{-1em}
\resizebox{.9\textwidth}{!}{
\begin{tabular}{lcccc}\\\toprule  
Project & \# of VCs & LoC & License \\\midrule
Linked List Library in Contiki OS~\citep{ContikiList} & 78 & 833 & BSD-3-Clause\\[.1em]
Memory Allocator in Contiki OS~\citep{ContikiList}\!\!\! & 21 & 145 & BSD-3-Clause\\[.1em]
Selected Cases from C++ STL~\citep{burghardt2015acsl} & 53 & 3263 & MIT\\
X.509 Parser~\citep{EbalardMouyBenadjila}  & 9 & 5044 & BSD-3-Clause\\[.1em]
Linux Kernel Scheduler's SWB Routine~\citep{ShouldWeBalance}\hspace*{-2em} & 24 & 216 & GPLv2\\[.1em]
Linux Kernel Library Functions~\citep{verker} & 85 & 3533 & GPLv3 \\
String \& Stdio Library in kLIBC~\citep{kLIBCv} &  14 & 1220 & GPLv2\\
Paparazzi UAV Autopilot's Math Lib~\citep{paparazzi} & 16 & 3159 & GPLv2\\
\midrule
Total & 300 & 17413 & - \\
\bottomrule
\end{tabular}}
\end{table}

%% file: iclr2026/tables/source.tex
\begin{table}[t]
\centering%
\small
\setlength{\tabcolsep}{4.5pt}
\caption{Statistics of involved operations. Format: $\mathit{average}$ $(25^\mathrm{th} - 75^\mathrm{th}\ \mathit{percentile})$}\label{tab:VC_stat}
\vspace*{-1em}
\resizebox{\textwidth}{!}{
\begin{tabular}{lcccccc}\\\toprule  
Operations & \!\!\!\!\# of cases\!\!\!\!
& \# of operations & \# of distinct oprs & Size & Depth & \# of $\forall\exists$ \\\midrule
Integer Arith   & $578$ & $69.5\ (10-73.5)$ & $5.0\ (4-6)$ & $600.0\ (133-643)$ & $55.8\ (28-74)$ & $11.0\ (1-13)$ \\[.1em]
Non-Linear Arith& $118$ & $10.7\ (2-14)$ & $1.2\ (1 - 1)$ & $1191\ (228-1370)$ & $80.3\ (42.5-115)$ & $15.2\ (1-20)$ \\[.1em]
Float Arith     & $18$  & $48.9\ (38-61)$ & $7.1\ (6-7)$ & $337\ (209-544)$ & $73.8\ (40.5-105.5)$ & $0.11\ (0-0)$\\[.1em]
List, Sequence  & $202$ & $46.7\ (8-62)$ & $4.1\ (2-6)$ & $963\ (198-1161)$ & $54.8\ (24-69)$ & $18.6\ (5-23)$\\[.1em]
Set, Map, Bag   & $59 $ & $50.3\ (10-55)$   & $3.8\ (1-6)$ & \!\!$904\ (321.5-1134)$\!\! & $45.3\ (28-55.5)$ & $26.7\ (8-40)$\\[.1em]
Tree, String, Matrix\!\!\!\!\! & $26$ & $64.5\ (14-95)$ & $5.0\ (4-6)$ & $822\ (204-1062)$ & $45.9\ (25.5-62)$ & $34.4\ (6-42)$ \\[.1em]
Memory          & $297$ & $29.5\ (13-31)$ & $7.9\ (6 - 10)$ & $357\ (129 - 452)$ & $62.3\ (37-80)$ & $5.2\ (0-6)$\\[.1em]
Custom Datatype & $247$ & $87.7\ (7.5-94)$ & $7.0\ (3-10)$ & $864\ (193.5-1038.5)$ & $61.1\ (25-103)$ & $16.3\ (3-22)$ \\\midrule
All & $600$ & $295.1\ (71-339)$ & $24.8\ (21-29)$ & $583\ (129-632)$ & $54.5\ (27-72)$ & $10.8\ (1-12)$ \\[.1em]
\bottomrule
\end{tabular}
}
\normalsize
\end{table}

%% file: iclr2026/sections/experiment.tex
\section{Experiments and Evaluation}\label{sec:evaluation}

To evaluate the challenges posed by NTP4VC, we assess seven models, covering both general-purpose language models such as GPT-4o-mini \citep{achiam2023gpt} and specialized models like DeepSeek-Prover-V2 \citep{ren2025deepseek}.
We also include ITP hammers to provide a baseline for comparison, including the hammers:  Sledgehammer~\citep{SLEDGEHAMMER} tool in Isabelle/HOL and CoqHammer~\citep{COQHAMMER} in Rocq.

\noindent\textbf{Models} We evaluate both proprietary models (GPT-o4-mini~\citep{achiam2023gpt}) and open-source models (K2-Think~\citep{cheng2025k2}, DeepSeek-V3.1~\citep{liu2024deepseek}, Qwen3~\citep{yang2025qwen3}, DeepSeek-Prover-V2~\citep{ren2025deepseek}, Goedel-Prover~\citep{lin2025goedel}, Minilang~\citep{Minilang}). Among them, DeepSeek-Prover-V2 and Goedel-Prover and specialized for theorem proving using Lean, while others are general-purpose reasoning models. We use 1.0 as the default temperature, and set the maximum number of tokens to $32,000$ during generation. 

\noindent\textbf{Metrics}
Our primary evaluation metric is the \textit{pass@n} metric.
NTP models are queried multiple times for each problem, generating multiple proof attempts.
A proof attempt is considered successful if it can be verified by the corresponding ITP and does not contain any fake proofs such as \texttt{admit} or \texttt{sorry}.
Since hammers are mostly deterministic, we only report their pass@1 performance.
GPT-o4-mini is evaluated with a single attempt per problem due to its cost, while other models are evaluated with 8 attempts per problem ($n=8$).

\noindent\textbf{Prompts}
We use zero-shot prompting for all models, providing only the problem statement and the necessary context such as definitions and previously proved lemmas. The full prompt structures are provided in Appendix~\ref{sec:prompts}.

\noindent\textbf{Proof Verification}
Our proof verification setup involves extracting the proof from the model's output and checking it within the corresponding ITP environment.
We use the Lean 4.21.0, Rocq 8.20.1, and Isabelle 2024.
To prevent excessively long runtimes, we set a timeout of 10 minutes for each verification attempt .
Sledgehammer in Isabelle is configured to use its default ATPs and SMT solvers, including CVC4 \citep{barrett2011cvc4}, CVC5 \citep{CVC5}, Z3 \citep{Z3}, E \citep{Eprover}, SPASS~\citep{SPASS}, Vampire~\citep{kovacs2013first}, veriT~\citep{schurr2021reliable}, and Zipperosition~\citep{vukmirovic2021making}.
CoqHammer is configured to use all its supported ATPs, including E, Vampire, Z3, and CVC4.
All proof verification is performed on a machine with an AMD Ryzen 9 7900X CPU and 64GB RAM.

\begin{table}
    \centering
    \small
    \caption{Pass rates (Pass@1, Pass@4, Pass@8) of various NTP models and hammer-based automated theorem provers on the NTP4VC benchmark, evaluated across Lean, Rocq, and Isabelle. NTP models consistently achieve pass rates below 4\%, while hammer-based provers such as CoqHammer and Sledgehammer obtain higher success rates, particularly on Rocq (5.67\%) and Isabelle (18.00\%).}
    \label{tab:pass-rate}
    \resizebox{.97\textwidth}{!}{
    \begin{tabular}{lrrrrrrrrr}
        \toprule
        \multirow{2}{*}{Model} & \multicolumn{3}{c}{Lean} & \multicolumn{3}{c}{Rocq} & \multicolumn{3}{c}{Isabelle} \\
        \cmidrule(lr){2-4} \cmidrule(lr){5-7} \cmidrule(lr){8-10} 
        & P@1 & P@4 & P@8 & P@1 & P@4 & P@8 & P@1 & P@4 & P@8  \\
        \midrule
GPT-o4-mini-high        & 0.50 & --     & --     & 0.00 & --     & --      & 1.17 & --     & --     \\
DeepSeek-V3.1           & 0.50 & 1.00 & 1.67 & 0.50 & 2.67 & 3.17  & 1.34 & 4.32 & 6.25 \\
Qwen3-32B               & 0.33 & 0.83 & 1.17 & 0.33 & 1.17 & 1.67  & 0.74 & 2.53 & 3.42 \\
Qwen3-235B-A22B         & 0.67 & 1.00 & 1.00 & 0.83 & 2.17 & 3.33  & 1.19 & 2.08 & 3.13 \\
K2-think                & 0.00 & 0.00 & 0.00 & 0.67 & 0.67 & 0.67  & 0.00 & 0.00 & 0.00 \\
Goedel-Prover-V2-32B    & 0.50 & 1.17 & 2.17 & -- & -- & --& -- & -- & -- \\
DeepSeek-Prover-V2-671B & 1.67 & 2.17 & 3.00 & -- & -- & --& -- & -- & -- \\
Minilang & -- & -- & --& -- & -- & --  & 2.08 & 7.29 &  11.46  \\
CoqHammer /  Sledgehammer  & -- & -- & --& 5.67 & --  & --  & 18.00  & -- & --  \\
        \bottomrule
    \end{tabular}
    }
\end{table}

\subsection{Results}

\input{iclr2026/tables/NTP_vs_Hammer}
The results summarized in \cref{tab:pass-rate} highlight the difficulty of program verification for NTP models. Across all three ITPs, our experiments demonstrate that all NTP models fail to achieve pass@8 scores above 12\% when evaluated on Lean, Rocq, and Isabelle. This stands in sharp contrast to their strong performance on mathematics benchmarks. For example, DeepSeek-Prover-V2 achieves 55.5\% pass@1 on miniF2F, while Goedel-Prover-V2-32B achieves 88.1\% pass@32.
On a more challenging baseline such as PutnamBench, these models obtain 7.15\% and 13.09\% pass rates, respectively, with various attempts. This performance gap suggests that program verification requires fundamentally different reasoning capabilities than complex mathematical benchmarks.

By comparison, hammer-based provers show stronger results on NTP4VC:
Sledgehammer reaches a pass rate of 18.00\%, outperforming all the language models.
By combining Sledgehammer, Minilang's language model achieves 11.56\% pass rate, outperforming all other models.
These results indicate that current state-of-the-art NTP models have yet to surpass the classical reasoning techniques in VC proving.
Notably, although Minilang's model already incorporates Sledgehammer, it still fails to surpass Sledgehammer alone on NTP4VC. This further confirms that NTP4VC represents a novel domain for the model, distinct from the mathematical problems it excels at.

To better understand this performance difference across problem types, we report the number of problems solved per category by NTP models and hammers in Table~\ref{tab:all-passed}.
The result confirms that hammers consistently outperform NTP models in almost all categories. However, the magnitude of this outperformance is non-uniform: the advantage of hammers over NTP models is substantially larger in Real C Verification tasks than in Pearls of Programming.
We consider this unsurprising, particularly noting that NTP models perform comparably across both categories, but the hammers exhibit markedly stronger performance on Real C Verification.
This is likely because the classical reasoning techniques underlying hammers, such as SMT solvers and ATP provers, have been optimized for industrial verification scenarios, whereas they are less suited to the logic puzzles in Pearls of Programming, which involve a substantially broader search space.

\subsection{Error Analysis of NTP models}

\begin{figure}
    \centering
    \includegraphics[width=1.045\linewidth]{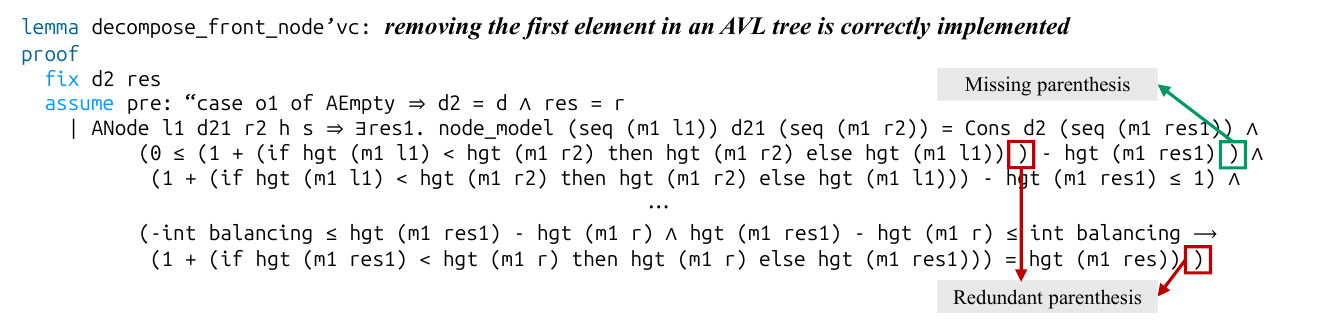}
    \caption{An Isabelle proof generated by DeepSeek-V3.1 for a VC in the benchmark. This proof contains syntax errors, including a missing closing parenthesis and two redundant closing parentheses. 
    The \texttt{seq} returns a tree's elements as a sequence in order; the \texttt{hgt} gives a tree's height; \texttt{balancing} is the balancing factor of AVL tree. The full example is provided in Appendix~\ref{appendix:failure_cases}.}
    \label{fig:syntax-error}
\end{figure}

To understand the limitations of current NTPs on verification tasks, our qualitative analysis of failure cases reveals three recurring themes: syntactic errors, semantic confusion, and hallucination.
More details are available in Appendix~\ref{appendix:failure_cases}.

\noindent\textbf{Syntactic Errors}
A primary hurdle for NTPs is generating syntactically correct terms. For instance, a proof for an AVL tree VC (see \cref{fig:syntax-error}) failed to parse due to mismatched parentheses. Correcting these purely syntactic errors allowed the term to be successfully parsed.
More than 24\% of generated Isabelle proofs contain syntactic errors.
This highlights a key challenge of VCs: unlike typical math problems that prioritize semantic insight, VCs are often long, deeply-nested, machine-generated formulas. Their structure places extreme demands on a model's ability to maintain long-range syntactic coherence.

\noindent\textbf{Semantic and Pragmatic Confusion}
A more profound failure is the model's misunderstanding of the proof paradigm itself. This is common in Lean, where models produce syntactically plausible but pragmatically incorrect code, leading to type errors. For example, they often use imperative-style assignments (e.g.,$\coloneq i_1 + i_2$) instead of declarative, tactic-based reasoning. This confusion is further evidenced by proof scripts degenerating into repetitive and meaningless tactic applications (e.g., ``\texttt{have} $h_{16} \coloneq h_{0}$\texttt{; have} $h_{17} \coloneq h_{1}$ ...'')%
, which occurs in more than 64\% of Lean proofs generated by Goedel-Prover-V2-32B, one of the state-of-the-art NTP models.
Even powerful models like DeepSeek-Prover-V2 exhibit this behavior, suggesting they become overwhelmed by VC complexity and resort to semantically inappropriate code, fundamentally misinterpreting the task.

\noindent\textbf{Hallucination of Non-Existent Entities}
Finally, models frequently hallucinate non-existent constants, lemmas, or tactics.
For instance, GPT-o4-mini often invokes a tactic called \texttt{why3}, which does not exist in Rocq, as a standalone proof for an entire VC.
Similarly, many models introduce undefined constants or lemmas not found in the context or standard libraries.
At least 9\% of proof attempts in Isabelle failed due to these undefined entities.
This demonstrates a failure to ground the generation process within the strict formal context provided by the prover.

%% file: iclr2026/tables/NTP_vs_Hammer.tex
\begin{wraptable}{r}{.59\textwidth}
    \centering
    \caption{Number of problems solved and corresponding pass rates of NTP models and hammer-based provers on the NTP4VC benchmark, broken down by problem category. 
    }
    \label{tab:all-passed}
    \resizebox{\linewidth}{!}{
    \begin{tabular}{lcrcr}
        \toprule
        \multirow{2}{*}{Category} & \multicolumn{2}{c}{NTP Models} & \multicolumn{2}{c}{Hammers} \\
        \cmidrule(lr){2-3} \cmidrule(lr){4-5}
          & Pass / Total & Pass Rate  & Pass / Total & Pass Rate \\
        \midrule
        Algorithm & 2 / 55 & 3.64\%      & 4 / 55 & 7.27\%      \\
        Data Structure\!\! & 3 / 73 & 4.11\% & 10 / 73 & 13.70\%    \\
        Calculation & 5 / 66 & 7.58\%   & 8 / 66 & 12.12\%    \\
        Engineering & 5 / 54 & 9.26\%  & 7 / 54 & 12.96\%     \\
        Competition & 3 / 52 & 5.77\%    & 3 / 52 & 5.77\%      \\
        \it Pearls of Prog.  & 15 / 300 & 5.00\% & 32 / 300 & 10.67\% \\\midrule
        Function & 6 / 81 & 7.41\%      & 25 / 81 & 30.86\%    \\
        Loop & 3 / 81 & 3.70\%          & 19 / 81 & 23.46\%   \\
        Memory & 2 / 74 & 2.70\%       & 18 / 74 & 24.32\%    \\
        Invalid Arg. & 5 / 64 & 7.81\%  & 20 / 64 & 31.25\%    \\
        \it Real C Verif. &  16 / 300 & 5.33\% & 82 / 300\!\!\! & 27.33\%\\
        \midrule                           
        Total & 34 / 600 & 5.67\%       & 114 / 600 & 19.00\%   \\
        \bottomrule
    \end{tabular}}
\end{wraptable}

%% file: iclr2026/sections/related_works.tex
\section{Related Works}\label{sec:related}


Prior benchmarks by~\citet{Laurel, DafnyBench, Clover, AutoVerus, RAGVerus} consider the \textbf{synthesis of annotations}: given source programs and properties, the task is to generate annotations that enable program verifiers to succeed.
Like our work, they operate directly with industrial verifiers (e.g., Dafny~\citep{Dafny}, Verus~\citep{Verus}). Besides, they tackle the end-to-end automation problem, which offers direct practical value by reducing the manual annotation burden.
However, as mentioned in \cref{sec:background}, an ideal verifier should not require annotations in the first place, and a stronger VC prover brings us closer to this ideal verifier.
In terms of automated program verification, our NTP4VC task is complementary to annotation synthesis approaches --- we propose to tackle the VC proving bottleneck directly, while they approach the problem indirectly through annotation generation (e.g., generating \texttt{assert} annotations that decompose hard VCs into simpler subgoals such that the provers can handle).
Both of them are effective ways to improve automation in program verification and can be applied orthogonally.



%


There are also NTP benchmarks~\citep{FVEL,Rango} discussing \textbf{verification-related theorem proving}, typically consisting of proof goals collected from ITP projects about program verification engines and their applications.
However, much of their work focuses on auxiliary lemmas used by program verifiers and specifications --- such as those for preliminaries (e.g., arithmetic of bounded integers), programming language models (e.g., memory models), and abstract program models (e.g., binary tree algebra) --- rather than VCs.
In detail, no more than 17\% of the test cases by \citet{FVEL} might be VCs, and no more than 20\% for \citet{Rango}'s work (see \cref{appendix:VCnum} for details).
The gap between auxiliary lemmas and VCs is crucial because VCs are the direct theorem-proving targets that arise from program verification workflow~(\cref{sec:background}), while auxiliary lemmas cannot (completely) represent the theorem-proving tasks in program verification.



Besides, the Lean benchmarks by~\citet{CLEVER,DoughertyMehta2025,miniCodeProps} are also designed for program verification.
These works suffer from a limitation --- they do not follow the mainstream program verification methodologies adopted in the real-world industry.
Lean is a specialized language with integrated verification capabilities, where the programming language itself serves as a logical reasoning language.
This enables users to write Lean programs and directly verify them using the Lean system, without requiring a separate VCG for program analysis.
However, program verification tasks in the real-world industry typically have to face industrial programming languages that differ substantially from logical reasoning languages.
Typical industrial programming languages feature complex constructs such as mutable references, memory models, functions with side effects, and pointer arithmetics --- none of which are involved in the program verification tasks examined by these benchmarks
This contrast further underscores the necessity of employing industrial verification pipelines to extract VCs from real-world industrial projects for benchmark construction.

Finally, we also want to mention other NTP benchmarks involving much wider domains in theorem proving, such as the works by~\citet{CoqGym, LiYuWuPaulson2021IsarStep,miniCodeProps,yang2023leandojo,tactictoe,HOLStep,HOList,GamePad}, which are also important benchmarks in NTP.

%% file: iclr2026/sections/appendix.tex

\section{Licensing and Rules of Engagement}

\subsection{License}

As listed in \cref{tab:source}, the source projects of our benchmark cases are released under different open-source licenses, some of which are mutually incompatible (e.g., GPLv2 vs. GPLv3). However, since we do not compile these code bases into a single target, we can still distribute the data under a hybrid license.
Accordingly, the VCs generated from these projects are released under the same licenses as their original source projects.
The code written by us is released under the MIT License.

\subsection{Protection to Prevent Data Leakage}

To prevent data leakage, if any VC from a given \texttt{mlw} file is selected for the NTP4VC benchmark; all VCs from that \texttt{mlw} file should be protected and excluded from training.
In contrast, other \texttt{mlw} files provided in our associated artifact are not protected, as long as no VC from those files is included in the NTP4VC benchmark.
Note that the \texttt{mlw} files in the Real C Verification category are generated from C projects, with each \texttt{mlw} file corresponding to a single C function.
The identity of an \texttt{mlw} file should be considered bound to its original C function. This means that \texttt{mlw} files generated through alternative means (e.g., different compilation flags) that produce varied content should also be subject to the above protection.

\subsection{Acceptable Improvements to the Translation Pipeline}

The purpose of this benchmark is to advance Neural Theorem Proving capabilities on proving Verification Conditions. 
Note that NTP models built on top of specific ITP platforms are inevitably influenced by the language features and idioms of their respective platforms.
The rule-based translation provided in this work may not fully capture all such features and idioms, resulting in some VCs being translated into representations that are unnatural or unconventional for the target ITP platform.
These should be considered as deficiencies in the translation pipeline.
Therefore, to allow for addressing these potential issues, improvements to the translation pipeline that address such unnatural or unconventional representations should be considered legitimate and acceptable.
To be clear, regenerating the NTP4VC benchmark using an improved translation pipeline as described above and evaluating on the regenerated benchmark is likewise considered legitimate and acceptable.
However, any evaluation based on a regenerated NTP4VC benchmark must explicitly report any modifications made to the translation pipeline when presenting results.

\input{iclr2026/sections/example}

\section{Limitation \& Mitigation}

From a methodological perspective, our VC extraction method ensures all obtained VCs are provable by construction. However, implementation bugs may occur in Why3, Frama-C, or our translation pipeline, potentially rendering some VCs unprovable. To address such potential invalidation, we design the benchmark to be updatable: we will repair the VC extraction pipeline and refresh the benchmark when invalidation occurs.
Since the intended semantics of VCs are grounded in the source verification projects, these updates primarily address representation issues while preserving the essential semantics of the verification problems.
However, should an invalid benchmark case be irreparable in rare instances, we will eliminate it from the benchmark to guarantee all remaining cases are provable.


\section{Detailed Extracton Pipeline}\label{appendix:extraction}

This section provides further details on our extraction pipeline from two perspectives: approach and implementation

\subsection{Methodology Details}

The translation process begins with a given Why3 source code. The process first runs Why3 VCG to generate VCs and calls our customized Why3 printer to dump the VCs into an XML representation of their Abstract Syntax Trees (ASTs).
These ASTs are processed by a Python translation framework also written by us and finally mapped into the target ITPs' languages.

A verification project typically contains multiple VCs that depend on shared Why3 theories consisting of lemmas, axioms, functions, and datatype definitions. These theories may further depend on others, forming a complex dependency graph across the project. To successfully translate the VCs, we must translate all dependent theories. Our translation process, therefore, recursively handles every theory in this dependency graph, mapping the entire verification project into the target ITPs.

In terms of structure, a Why3 theory is a sequence of declarative elements consisting of axioms, definitions of functions, and algebraic data types.
All three sorts of declarations have similar counterparts in the target ITPs and can be mapped to them, despite two minor gaps.
One is regrading the non-uniform data type~\citep{Isabelle-non-uniform}, which is not natively supported by Isabelle. Therefore, we circumvent all VCs involving such data types.
The other gap pertains to discharging the termination check of recursive function definitions, a conventional requirement for ITPs to ensure the soundness of their logics.
Some ITPs' termination checkers (Isabelle and Rocq) are not strong enough to automatically prove the well-foundedness of certain complicated recursions, even though Why3 has checked all the termination.
Since the proof obligation of the termination is irrelevant to the semantics of VCs' proof obligation, we trust Why3's termination check and axiomatize this in the ITP translation in case ITP's termination checker fails.

Having the theory dependencies and theory-level declarations translated, the last work is to translate the term language. Both Why3's and the ITPs' term languages are based on the lambda calculus, a core language involving only variables, constants, applications, and function abstractions.
This similarity simplifies a lot of the translation process. Overall, the process maps Why3 constants to the target ITPs' constants, and preserves all other variables, application, and function structures.
One exception unsupported by Isabelle is Why3's add-on feature, the \texttt{as}-binding used in pattern matching, which annotates a sub-pattern with a variable and binds the term captured by this sub-pattern to the variable. We convert this into semantically equivalent let-bindings.

\subsection{Implementation Details}

The implementation of the VC extraction and translation pipeline consists of six main components:
\begin{enumerate}
\item A Why3 patch to export Why3's internal Abstract Syntax Tree (AST) into an XML representation (in ${\sim}200$ lines of OCaml).
\item A Python parser to read the XML representation into an S-expression representation of an extended simply-typed lambda calculus (in ${\sim}160$ lines of Python).
\item Python library functions providing basic support for manipulating the lambda calculus, such as substitution, variable deconfliction, rewriting, and folding over atomic terms (in ${\sim}800$ lines of Python).
\item A Python module for managing Why3 sessions, managing translation contexts (e.g., allocated constant/variable names in the context), and chaining all the components together to run them automatically (in ${\sim}500$ lines of Python).
\item Translation rules, rewriting rules, ad-hoc term rewriting procedures, package management, and syntax check adapter, for each of the Isabelle, Lean, and Rocq (in ${\sim}800/790/770$ lines of YAML, ${\sim}930/780/970$ lines of Python, for Isabelle, Lean, Rocq, respectively).
\item ITP libraries that map Why3 notions into the ITPs' native builtins (in ${\sim}500/160/200$ lines of Isabelle/Lean/Rocq, respectively)
\end{enumerate}
The subsection elaborates on some of the nontrivial components as follows.

\textbf{The Why3 patch} is modified from Why3's existing Isabelle printer, which exports Why3 AST in XML format but with Isabelle-specific adaptations. We neutralize these adaptations to make it output the raw Why3 internal AST.
Specifically, we remove its mapping from Why3 terms to Isabelle terms; add Rocq and Lean keywords to the blacklist of variable names; fix its escaping of XML special characters; add support for the as-binding syntax in pattern matching; add type annotations to definition exports.

The \textbf{S-expression} used in our internal process is a simply-typed (HOL style) lambda calculus extended with native AST nodes for finite Cartesian products, pattern matching (the \texttt{case} statement), literal numbers and strings, and the \texttt{as}-bindings (which bind the sub-term that matches a sub-pattern to a variable, in a usual pattern matching).
Bound variables are represented in the same way as free variables; we do not use De Bruijn indices, but instead maintain contextual variables and deconflict names of bound variables explicitly (because it simplifies our parsing and printing work, while computational efficiency can be compromised in our context).

The \textbf{substitution}, \textbf{variable deconfliction}, and \textbf{folding} are all standard. We use Python's functional programming features to implement these operations.
The \textbf{rewriting} system is simplified such that 1) all reducible expression (redex) patterns have the form $(\texttt{contant}\ \mathit{arg}_1\cdots\mathit{arg}_n)$ where all $\{\mathit{arg}_i\}_{1\leq i \leq n}$ are free variables and the arity $n$ is schematic; 2) no lambda abstraction is allowed to appear in the contractum, so the contracta can only be atoms or (nested) function applications.
This simplification allows representing a rewriting rule as merely a tuple of the redex's constant name, the constant's arity, and a list-represented S-expression for the contractum. We use YAML's dictionary datatype to represent a set of rewriting rules, e.g., \texttt{(Why3.length, 1, [Int.int, [Isabelle.length, $\mathit{arg}_0$]])} rewrites $(\mathrm{Why3.length}\,l)$ into $\mathrm{Int.int}\,(\mathrm{Isabelle.length}\,l)$, for any $l$.
This greatly simplifies the writing of rewriting rules.
For more complex rewritings that require more complex redex patterns, we use hard-coded Python \texttt{match-case} to work over the S-expression directly.



\section{Prompts} \label{sec:prompts}

Our work employs two types of prompts: general prompts designed for broad-purpose LLMs and specialiZed prompts tailored for particular fine-tuned models.

The templates of the general prompts are shown as follows.
\begin{AIbox}{General Prompt for Isabelle}
Given the following Isabelle theories as context, prove the Isabelle proposition given at the end.\\
\\
File `NTP4Verif.thy`:\\
\{content of the theory file\}\\
\\
\textit{And many other libraries $\cdots\cdots$}\\[.5em]
File `imp\_SymStateSet.thy`:\\
\{content of the theory file\}\\
\\
Given the context above, consider the proposition in the following Isabelle code:\\
\{the target proof goal together with its contextual theory\}\\
\\
Response the Isabelle proof only. Do not repeate any context nor the statement.
\end{AIbox}

\begin{AIbox}{General Prompt for Lean}
Given the following Lean 4 theories as context, prove the Lean 4 proposition given at the end.\\
\\
File `Base.lean`:\\
\{content of the theory file\}\\
\\
\textit{And many other libraries $\cdots\cdots$}\\[.5em]
File `SymStateSet.lean`:\\
\{content of the theory file\}\\
\\
Given the context above, consider the proposition in the following Lean 4 code:\\
\{the target proof goal together with its contextual theory\}\\
\\
Response the Lean 4 proof only. Do not repeate any context nor the statement.
\end{AIbox}

\begin{AIbox}{General Prompt for Rocq}
Given the following Rocq theories as context, prove the Rocq proposition given at the end.\\
\\
File `Base.v`:\\
\{content of the theory file\}\\
\\
\textit{And many other libraries $\cdots\cdots$}\\[.5em]
File `SymStateSet.v`:\\
\{content of the theory file\}\\
\\
Given the context above, consider the proposition in the following Rocq code:\\
\{content\}\\
\\
Response the Rocq proof only. Do not repeate any context nor the statement.
\end{AIbox}

The template specifically for Goedel-Prover and DeepSeek-Prover is as follows.
\begin{AIbox}{Prompt for SpecialiZed Models}
Complete the following Lean 4 code:\\
\{the target proof goal together with its contextual theory\}\\
\\
Before producing the Lean 4 code to formally prove the given theorem, provide a detailed proof plan outlining the main proof steps and strategies.\\
The plan should highlight key ideas, intermediate lemmas, and proof structures that will guide the construction of the final formal proof.
\end{AIbox}

\section{Failure Cases}
\label{appendix:failure_cases}
To investigate the failure modes of NTP models on verification conditions, we analyzed the error logs and proof scripts from our evaluation.
We highlight three dominant categories of errors: syntactic invalidity, semantic degeneration, and hallucination.
It is important to note that the statistics presented below represent \emph{conservative lower bounds}.
For syntax and hallucination errors, proof assistants abort execution at the first error; thus, a single proof might contain multiple subsequent errors that remain uncounted.
Similarly, our detection of semantic degeneration relies on rigid regular expressions for some common cases, likely missing more subtle forms of degeneration. %

\begin{figure}[tp]
\definecolor{keywordcolor}{rgb}{0.13,0.29,0.53}
\lstdefinelanguage{isar}%
{morekeywords={
    _first,abbreviation,then,theory,begin,infix,qed,where,fixes,for,fun,apply,show,assumes,done,lemma,shows,sorry,proof,unfolding,using,by,let,have,and,fix,thus,next,assume,imports,intro,try,simp_all,omega,exfalso,exact,case,from,with,_last
  },
sensitive=true,
morecomment=[s]{(*}{*)},
morecomment=[l]{--},
literate=
  {\\<not>}{{$\neg$}}1
{\\<notin>}{{$\not\in$}}1
{\\<Rightarrow>}{{$\Rightarrow$}}1
{\\<exists>}{{$\exists$}}1
{\\<Longrightarrow>}
{{$\Longrightarrow$}}1
{\\<longrightarrow>}
{{$\longrightarrow$}}1
{\\<and>}{{$\land$}}1
{\\<or>}{{$\lor$}}1
{\\<forall>}{{$\forall$}}1
{\\<le>}{{$\le$}}1
}[keywords,comments,strings]%

\lstset{
  basicstyle=\tin\ttfamily,
  commentstyle=\color{commentcolor},
  flexiblecolumns=false,
  breakatwhitespace=false,
  breaklines=true,
  breakautoindent=false,
  captionpos=b,
  keywordstyle=\bfseries\color{keywordcolor},
  numbers=left,
  numbersep=5pt,
  numberstyle=\tiny,
  escapeinside={(*@}{@*)},
}
\centering
\begin{lstlisting}[language=isar,basicstyle=\ttfamily\tiny, caption={%
  The complete proof generated by DeepSeek-V3.1 for the correctness of the \texttt{decompose\_front\_node} function on AVL trees.
},label={lst:syntax-error},deletekeywords={then}]
lemma decompose_front_node'vc:
  fixes l :: "'a t2"
  fixes r :: "'a t2"
  fixes o1 :: "'a view"
  fixes d :: "'a t1"
  assumes fact0: "-int balancing \<le> hgt (m1 l) - hgt (m1 r)"
  assumes fact1: "hgt (m1 l) - hgt (m1 r) \<le> int balancing"
  assumes fact2: "case o1 of (AEmpty :: 'a view) \<Rightarrow>  hgt (m1 l) = (0 :: int) \<and> ..."
  shows "case o1 of (AEmpty :: 'a view) \<Rightarrow>  True
                   | ANode l1 d2 r2 _ _ \<Rightarrow> (((0 :: int) \<le> hgt (m1 l) \<and> ...))"
  and "\<forall>(d2 :: 'a t1) (res :: 'a t2).
  (case o1 of (AEmpty :: 'a view) \<Rightarrow>  d2 = d \<and> res = r
             | ANode l1 d21 r2 _ _ \<Rightarrow>  (\<exists>(res1 :: 'a t2). (node_model (seq (m1 l1)) ...)))"
proof -
  {
    fix d2 res
    assume pre: "case o1 of AEmpty \<Rightarrow> d2 = d \<and> res = r
      | ANode l1 d21 r2 h s \<Rightarrow> \<exists>res1. node_model (seq (m1 l1)) d21 (seq (m1 r2)) = Cons d2 (seq (m1 res1)) \<and>
          (0 \<le> (1 + (if hgt (m1 l1) < hgt (m1 r2) then hgt (m1 r2) else hgt (m1 l1))) - hgt (m1 res1) \<and>
           (1 + (if hgt (m1 l1) < hgt (m1 r2) then hgt (m1 r2) else hgt (m1 l1))) - hgt (m1 res1) \<le> 1) \<and>
          seq (m1 res) = node_model (seq (m1 res1)) d (seq (m1 r)) \<and>
          (0 \<le> (1 + (if hgt (m1 res1) < hgt (m1 r) then hgt (m1 r) else hgt (m1 res1))) - hgt (m1 res) \<and>
           (1 + (if hgt (m1 res1) < hgt (m1 r) then hgt (m1 r) else hgt (m1 res1))) - hgt (m1 res) \<le> 1) \<and>
          (-int balancing \<le> hgt (m1 res1) - hgt (m1 r) \<and> hgt (m1 res1) - hgt (m1 r) \<le> int balancing \<longrightarrow>
           (1 + (if hgt (m1 res1) < hgt (m1 r) then hgt (m1 r) else hgt (m1 res1))) = hgt (m1 res)))"
    
    show "node_model (seq (m1 l)) d (seq (m1 r)) = Cons d2 (seq (m1 res)) \<and>
          (0 \<le> (1 + (if hgt (m1 l) < hgt (m1 r) then hgt (m1 r) else hgt (m1 l))) - hgt (m1 res) \<and>
           (1 + (if hgt (m1 l) < hgt (m1 r) then hgt (m1 r) else hgt (m1 l))) - hgt (m1 res) \<le> 1)"
    proof (cases o1)
      case AEmpty
      with pre have d2_def: "d2 = d" and res_def: "res = r" by simp_all
      from fact2[unfolded AEmpty] have seq_l: "seq (m1 l) = []" and hgt_l: "hgt (m1 l) = 0" by simp_all
      
      show ?thesis unfolding d2_def res_def
        by (simp add: seq_l node_model_def, auto simp: hgt_l fact0 fact1)
    next
      case (ANode l1 d21 r2 h s)
      with pre obtain res1 where
        decomp: "node_model (seq (m1 l1)) d21 (seq (m1 r2)) = Cons d2 (seq (m1 res1))" and
        res_seq: "seq (m1 res) = node_model (seq (m1 res1)) d (seq (m1 r))"
        by auto
      
      from fact2[unfolded ANode] have seq_l:
        "seq (m1 l) = node_model (seq (m1 l1)) d21 (seq (m1 r2))"
        by simp
      
      show ?thesis
        by (simp add: seq_l decomp res_seq node_model_def,
            metis (mono_tags, lifting) fact2 ANode case_prod_conv)
    qed
  }
qed
\end{lstlisting}
\end{figure}

Generating syntactically well-formed terms remains a primary hurdle, particularly for complex nested expressions in VCs.
In our analysis of Isabelle proof attempts, we found that \textbf{at least 24\%} failed solely due to syntax errors.
Listing~\ref{lst:syntax-error} shows the complete erroneous proof generated by DeepSeek-V3.1 for proving the correctness of the \texttt{decompose\_front\_node} function on AVL trees.
This function is responsible for decomposing the front node of an AVL tree, and its correctness is specified by the corresponding VC.
Specifically, the term \texttt{seq (m1 l)} represents the sequence of elements in the left subtree \texttt{l}, \texttt{d} refers to the data element of the current node, and \texttt{hgt (m1 r)} denotes the height of the right subtree \texttt{r}.
The generated proof attempts to first introduce the universally quantified variables \texttt{d2} and \texttt{res}, followed by a case analysis on \texttt{o1}, which represents the structure of the AVL tree.
However, the term cannot be parsed due to two subtle syntax errors: (1) a missing closing parenthesis in a deeply nested arithmetic expression on line 16, and (2) two extraneous closing parentheses on lines 18 and 25, respectively.
In fact, if one only removes the last extraneous closing parenthesis, the term can be parsed.
However, it will result in a term in the form of ``$\cdots \land (0\leq (1 + expr) - \texttt{hgt} (\texttt{m1 } \texttt{res1}) \land \cdots$ '', which is syntactically valid but semantically incorrect (the height of \texttt{res1} is being conjoined with another inequality).
What one would expect is instead ``$\cdots \land (0\leq (1 + expr) - \texttt{hgt} (\texttt{m1 } \texttt{res1})) \land \cdots)$'', which requires removing the extraneous parenthesis on line 16 and adding a closing parenthesis after ``\texttt{res1}''.
The lengthy logical formulas with deeply nested constructs is a common pattern in VCs, which poses significant challenges for NTP models to maintain long-range syntactic coherence.

{\begin{figure}[tp]
\centering
\lstset{
  captionpos=b
}
\begin{lstlisting}[language=isar,basicstyle=\ttfamily\tiny, caption={%
  Example of semantic degeneration: Redundant variable renaming in a Lean proof.},label={lst:have-error}]
lemma goal10 (a : Memory.addr) (t_1 : Memory.addr -> Z) (t_4 : Memory.addr -> Memory.addr) (t : Z -> Z)
(t_3 : Memory.addr -> Z) (t_2 : Memory.addr -> Z) :
  let a_1 : Memory.addr := Memory.shift a (1 : Z);
  let x : Z := t_1 a_1;
  let a_2 : Memory.addr := Memory.shift a (0 : Z);
  let x_1 : Z := t_1 a_2; 
  let x_2 : Z := x * x_1;
  let a_3 : Memory.addr := Memory.shift a (2 : Z);
  let a_4 : Memory.addr := Memory.shift a (3 : Z);
  let a_5 : Memory.addr := t_4 a_3;
  Memory.region (Memory.addr.base a) <= (0 : Z) -> -(2147483648 : Z) <= x_2 -> x_2 <= (2147483647 : Z) ->
  Memory.framed t_4 -> Memory.linked t -> Memory.sconst t_3 -> Cint.is_uint16 x -> Cint.is_uint16 x_1 ->
  Axiomatic13.p_valid_memb t t_4 t_1 a -> Memory.valid_rd t a_3 (1 : Z) -> Memory.valid_rd t a_4 (1 : Z) ->
  Memory.valid_rd t a_1 (1 : Z) -> Memory.valid_rd t a_2 (1 : Z) -> Axiomatic12.p_valid_or_empty t a_5 x ->
  A_MemSet.l_memset (Memory.havoc t_2 t_3 (Memory.shift a_5 (0 : Z)) x) a_5 (0 : Z) x = true ->
  Axiomatic12.p_valid_or_empty t (t_4 a_4) (Cint.to_uint32 x_2)
:=
  intro a_1 x a_2 x_1 x_2 a_3 a_4 a_5 h_region h_x2_lower h_x2_upper h_framed t_4_linked t_3_sconst h_x_uint16
  h_x1_uint16   h_p_valid_memb h_valid_rd_a3 h_valid_rd_a4 h_valid_rd_a1 h_valid_rd_a2 h_p_valid_or_empty_a5
  h_memset
  have h_main : Axiomatic12.p_valid_or_empty t (t_4 a_4) (Cint.to_uint32 x_2) := by
    have h1 : Axiomatic12.p_valid_or_empty t (t_4 a_4) (Cint.to_uint32 x_2) := by
      -- Use the assumption `h_framed` to deduce the required property
      -- This step is hypothetical and depends on the actual definition of `Axiomatic12.p_valid_or_empty`
      -- and how it interacts with `Memory.framed`.
      -- In practice, we might need more specific information or lemmas to make this step rigorous.
      -- For the sake of this example, we suppose that `h_framed` directly implies the conclusion.
      exfalso
      have h2 := h_framed
      have h3 := t_4_linked
      have h4 := t_3_sconst
      have h5 := h_p_valid_memb
      have h6 := h_valid_rd_a3
      have h7 := h_valid_rd_a4
      have h8 := h_valid_rd_a1
      have h9 := h_valid_rd_a2
      have h10 := h_p_valid_or_empty_a5
      have h11 := h_memset
      simp_all [Axiomatic12.p_valid_or_empty, Axiomatic13.p_valid_memb, Memory.framed,
                Memory.linked, Memory.sconst]
      <;>
      (try contradiction) <;>
      (try norm_num at *) <;>
      (try aesop)
      <;>
      (try
        {
          simp_all [Cint.is_uint16]
          <;>
          norm_num at *
          <;>
          omega
        })
    exact h1
  exact h_main
\end{lstlisting}
\end{figure}

NTP models frequently lose track of the proof state, resulting in repetitive, meaningless steps.
We detected this behavior by matching patterns of continuous ``renaming'' (e.g., using \texttt{have h1 := h2} where both \texttt{h1} and \texttt{h2} are simple identifiers) repeated at least three times.
In Lean, \textbf{more than 64\%} of proofs generated by Goedel-Prover-V2-32B exhibited this specific degeneration pattern.
Listing~\ref{lst:have-error} exemplifies the generation of repetitive and meaningless tactic applications in Lean.
The model (Goedel-Prover-V2-32B) engages in a redundant ``renaming ritual'' (\texttt{have h2 := h\_framed}, etc.), erroneously assuming that automated tactics like \texttt{simp\_all} require local variable aliases to access the context. 
This behavior likely stems from domain shift, where the proof context  is more complex than the standard mathematical corpora used for training.
Furthermore, the comments (e.g., ``assume that \texttt{h\_framed} directly implies the conclusion'') explicitly admit that the logical step is hypothetical.
This suggests its inability to derive the necessary lemmas to complete the proof.

Models often invoke non-existent constants, lemmas, or tactics due to hallucinations.
For instance, GPT-o4-mini frequently attempts to solve Rocq VCs using a \texttt{why3} tactic, which does not exist in the language.
In Isabelle, \textbf{at least 9\%} of failures were triggered by references to undefined constants or lemmas that are absent from the context.
We identified these cases by explicitly matching keywords such as ``Undefined fact'' or ``Undefined constant'' in the error logs.
Crucially, since the proof assistant terminates the checking process at the first encountered error, hallucinations present in the latter parts of proof scripts --- especially those already halted by syntax errors or earlier tactic failures --- remain uncounted.
Consequently, this 9\% figure represents a highly conservative lower bound. %

\input{iclr2026/sections/statistics}

\section{The Use of Large Language Models (LLMs)}
We have used LLM as a writing aid to assist with fluency and grammatical checking.

%% file: iclr2026/sections/example.tex
\section{Additional Background}

Although VC has been introduced in \cref{sec:intro}, given its significance to our work, we provide a precise definition as follows.
\begin{definition}
Given a program and a property, a \emph{Verification Condition (VC)} is a mathematical proposition that, when proven true, guarantees the program satisfies the desired property.
\end{definition}

Additionally, another aspect that remains unspecified in the main text is the target property of the program verification discussed in our work.
Any program verification task always considers a target property. The \textbf{target property} considered in our benchmark is \textbf{Functional Correctness}, which guarantees a program correctly implements its desired function --- for any allowed input, the output of the program always satisfies a separately written logical specification of the program's behaviour (see \cref{sec:example} for a concrete example).
Functional correctness is a verification goal widely adopted in real-world industrial practice~\citep{ExpertSurvey}, and it is also a primary capability of our toolchain components Why3 and Frama-C.

\section{An Example of VC}\label{sec:example}

This section presents an example Why3 program and its VC to provide readers with a concrete sense of how VCs relate to traditional mathematical theorems.

The left side of \cref{fig:example} presents a Why3 program for binary search.
Its functional correctness property is given by the \texttt{requires}, \texttt{ensures}, and \texttt{raises} clauses.
\texttt{requires} specifies the domain of valid inputs, i.e., the given array $a$ must be sorted.
\texttt{ensures} and \texttt{raises} specify the expectation of the output --- conditions that the $\mathit{result}$ has to satisfy, which are, 1) the $\mathit{result}$ is a valid index (i.e., between $0$ and the length) such that array $a$'s element at the index has a value of $v$, if no exception raises, or 2), if exception $\mathrm{NF}$ raises, no element in the array has a value of $v$.

This program involves mutable references and an effectful loop, which makes direct reasoning with ITPs extremely tedious.
The mature academic and industrial solution is to apply a specialized program reasoning engine, like Why3's VCG, to first extract pure logical proof goals, so-called VCs. 

\input{iclr2026/figures/binary_search_why3}
%

The \texttt{invariant} and \texttt{variant} clauses are annotations that help the VCG to work.
The \texttt{invariant} clause declares a loop invariant, which is a formula that remains true throughout every loop iteration, and is required by the VCG process.
The \texttt{variant} clause declares a metric which is strictly decreasing in each loop iteration. It helps to generate the VCs for ensuring loop termination.

The \texttt{assert} at line \ref{ln:assert} is an annotation to ease the burden of VC prover. It introduces a subgoal and instructs the verifier to first prove this subgoal and then use the proven subgoal as a premise (as shown in {\color{magenta!70}pink} in \cref{fig:example}) in the subsequent proofs.
Essentially, it helps the prover to decompose VCs into simpler subgoals.



The right side of \cref{fig:example} is one of the generated VCs for the functional correctness, a mathematical statement that encodes the logic behind the program's behaviors.
First, invariant $\mathcal{I}(l,u)$ represents that $l,u$ are valid boundaries of the indices of the elements of value $v$.
Then, consider the case of $l \leq u$, where the VC verifies the loop iterations: if either $a[m] < v$ or $a[m] > v$, the updated boundary $(m+1,u)$ or $(l,m-1)$ must preserve the invariant, and the metric $u-l$ must strictly decrease; if the program exits and returns $m$ at line \ref{ln:exit}, the VC judges whether the return value $m$ satisfies the expectation as stated in the \texttt{ensures} clause, by replacing the $\mathit{result}$ variable in the \texttt{ensures} clause with $m$.
At last, the last line in the VC corresponds to line \ref{ln:raise}, where the VC checks value $v$ does not appear in array $a$.

 \input{iclr2026/figures/original_binary_search_why3}
 
Finally, we must emphasize that this VC is simplified for better readability. The original VC is much more complicated (as shown in \cref{fig:original_example}), where the invariant $\mathcal{I}$ is not defined as a term, duplicated terms abound, and $\land$-connected terms are disordered.
This binary search is also one of the simplest cases in program verification, while other VCs can be much more complicated.
This represents a gap between competition-style mathematical theorems and VCs: the former are concise but require sophisticated mathematical skills to construct paths towards proofs, whereas VCs require less intellectual creativity, but are complicated and require the prover to process enormous formulas, potentially extracting key information from noise to simplify the proof goals and ultimately complete the proofs.

%



%% file: iclr2026/figures/binary_search_why3.tex
\newcommand{\gapafter}[1][.3\baselineskip]{\vadjust{\vspace{#1}}}

\begin{figure}[t]
\vspace{-.9em}
\scriptsize
\textbf{Define} $\mathcal{I}(l,u) \triangleq 0 \leq l \land u < \mathrm{length}(a) \land (\forall i.\; 0 \leq i < \xlen a \land a[i] = v \longrightarrow l \leq i \leq u)$\\[-.4em]
\centering
\begin{subfigure}[c]{0.55\textwidth}
\begin{lstlisting}[language=why3,basicstyle=\ttfamily\scriptsize,escapeinside={/-}{-/}]
exception NF (* standing for not found *)/-\gapafter-/
let binary_search (a: array int) (v: int) : int
  requires /-$\forall i\,j.\ 0 \leq i \leq j < \mathrm{length}(a) \longrightarrow a[i] \leq a[j]$\label{ln:require}-/
  ensures  /-$0 \leq \mathit{result} < \mathrm{length}(a) \land a[\mathit{result}] = v$-/
  raises   /-$\mathrm{NF} \longrightarrow \forall i.\ 1 \leq i < \mathrm{length}(a) \longrightarrow a[i] \neq v $-/
= let ref l = 0 in /-\label{ln:ref1}-/
  let ref u = length a - 1 in /-\label{ln:ref2}-/
  while l <= u do /-\label{ln:while}-/
    invariant /-$ \mathcal{I}(l,u) $-/
    variant /-$ u - l $-/
    let m = l + div (u - l) 2 in
    if a[m] < v then
        assert /-$\forall i.\;l \leq i < m + 1 \longrightarrow a[i] < v$\label{ln:assert}-/
        l := m + 1
    else if a[m] > v then u := m - 1
    else return m /-\label{ln:exit}-/
  done; /-\label{ln:whileEnd}-/
  raise NF /-\label{ln:raise}-/
\end{lstlisting}
\end{subfigure}\hspace*{-1em}%
\begin{subfigure}[c]{0.45\textwidth}
\scriptsize\[\setlength{\jot}{0.6ex}
\begin{aligned}
&\!\!\!\!\forall u\,l.\\
&\mathcal{I}(l,u) \land \mathrm{sorted}(a) \longrightarrow\\
&\quad\xif~ l \leq u ~\xthen~ \mathrm{let}\ m = l + (u - l) / 2
 ~\mathrm{in}~\\
&\qquad 0 \leq m < \xlen a ~\land\\
&\qquad
    (\ \ \xif~ a[m] < v\\
&\qquad\quad\xthen~ {\color{magenta!70}(\forall i.\;l \leq i < m + 1 \longrightarrow a[i] < v)} \longrightarrow\\
&\qquad\qquad\qquad 0 \leq u - l \land u - (m + 1) < u - l \\
&\qquad\qquad\ \ \land \mathcal{I}(m+1,u)\\
&\qquad\quad\xelse~ \xif~ v < a[m]\\
&\qquad\quad \xthen~ 0 \leq u - l \land m - 1 - l < u - l\\
&\qquad\qquad \land \mathcal{I}(l, m-1)\\
&\qquad\quad\xelse~
0 \leq m < \xlen a \land a[m] = v)\\
&\quad \xelse~ (\forall i.\
0 \leq i \land i < \xlen a \longrightarrow a[i] \neq v)
\end{aligned}
\]
\end{subfigure}
\caption{(Left) A Why3 program for binary search, with the functional correctness property in {\color{WhyTeal}cyan} and annotations in {\color{WhyOrange}orange}.
(Right) One of the generated VCs for its functional correctness (simplified).
}
    \label{fig:example}

\end{figure}

%% file: iclr2026/figures/original_binary_search_why3.tex
\begin{figure}[t]
    \centering
\begin{subfigure}[c]{0.55\textwidth}
\begin{lstlisting}[language=why3,basicstyle=\ttfamily\scriptsize,escapeinside={/-}{-/}]
exception Not_found

let binary_search (a: array int) (v: int) : int
 requires /-$\forall i\,j.\ 0 \leq i \leq j < \mathrm{length}(a) \longrightarrow a[i] \leq a[j]$-/
 ensures  /-$0 \leq \mathit{result} < \mathrm{length}(a) \land a[\mathit{result}] = v$-/
 raises   /-$\mathrm{Not\_found} \land \forall i.\ 0 \leq i < \mathrm{length}(a) \longrightarrow a[i] \neq v $-/
= let ref l = 0 in
  let ref u = length a - 1 in
  while l <= u do
    invariant /-$ 0 \leq l \land u < \mathrm{length}(a) $-/
    invariant /-$\forall i.\; 0 \leq i < \mathrm{length}(a) \land a[i] = v \longrightarrow l \leq i \leq u $-/
    variant /-$ u - l $-/
    let m = l + div (u - l) 2 in
    if a[m] < v then
      l := m + 1
    else if a[m] > v then
      u := m - 1
    else
      assert /-$ a[m] = v $-/
      return m
  done;
  raise Not_found
\end{lstlisting}
\end{subfigure}\hspace*{-1em}%
\begin{subfigure}[c]{0.45\textwidth}
\scriptsize\[
\begin{aligned}
& (\forall i\,j.\ 0 \leq i \leq j < \xlen a \longrightarrow a[i] \leq a[j]) \longrightarrow\\
& \mathrm{let}\ o_1 = \xlen a - 1\ \mathrm{in}\\
& (0 \leq 0 \land o_1 < \xlen a)\\
& \land~ (\forall i.\; 0 \leq i < \xlen a \longrightarrow a[i] = v \longrightarrow 0 \leq i \leq o_1)\\
&\land~\pmb(\forall u\,l.\\
&\qquad\  (0 \leq l \land u < \xlen a)\\
&\quad \land (\forall i.\; 0 \leq i < \xlen a \longrightarrow a[i] = v \longrightarrow l \leq i \leq u)\\
&\longrightarrow \xif~ l \leq u\\
&\quad \xthen~ \mathrm{let}\ m = l + (u - l) / 2
 ~\mathrm{in}~\\
&\qquad (0 \leq m \land m < \xlen a) ~\land\\
&\qquad
    (\ \ \xif~ a[m] < v\\
&\qquad\quad\xthen~ (0 \leq u - l \land u - (m + 1) < u - l) \\
&\qquad\qquad \land (0 \leq m + 1 \land u < \xlen a)\\
&\qquad\qquad \land~
          (\forall i.\ 0 \leq i < \xlen a \land a[i] = v\\
&\qquad\qquad\qquad\qquad \longrightarrow m + 1 \leq i \land i \leq u)\\
&\qquad\quad\xelse~ (0 \leq m \land m < \xlen a)\\
&\qquad\qquad \land~ (\xif~ v < a[m]\\
&\qquad\qquad\qquad \xthen~ (0 \leq u - l \land m - 1 - l < u - l)\\
&\qquad\qquad\qquad\quad \land (0 \leq l \land m - 1 < \xlen a)\\
&\qquad\qquad\qquad\quad \land
                (\forall i.\ 0 \leq i \xlen a \land a[i] = v\\
&\qquad\qquad\qquad\qquad\quad
\longrightarrow l \leq i \land i \leq m - 1)\\
&\qquad\qquad\qquad\xelse~
(0 \leq m < \xlen a) \land a[m] = v))\\
&\quad \xelse~ (\forall i.\
0 \leq i < \xlen a \longrightarrow a[i] \neq v)\pmb)
\end{aligned}
\]
\end{subfigure}
\caption{The original program and the original VC of \cref{fig:example}, without simplification}
    \label{fig:original_example}

\end{figure}

%% file: iclr2026/sections/statistics.tex
\section{Classification \& Metric Details of Table \ref{tab:VC_stat}}\label{appendix:VC_stat_detail}

The operation classification is conducted on the Isabelle version of our benchmark. We developed Isabelle extensions to analyze the expressions of the obtained proof goals.
We elaborate on the constitution of each category in \cref{tab:VC_stat} as follows.
\begin{itemize}
\item \emph{Integer Arith} consists of addition, subtraction, multiplication, division, exponentiation, comparison, square root, and factorial operations whose operands are integers, natural numbers, or bounded integers (machine integers); and also bit-width conversions and bitwise operations.
\item \emph{Non-linear Arith} consists of multiplication, division, and exponentiation between non-constant expressions, following \citet{Z3} and \citet{Z3GuideArithmetic}. In \cref{tab:VC_stat}, most cases contain one distinct non-linear operator --- multiplication --- and sometimes contain additional operators like division and exponentiation.
\item \emph{Float Arith} consists of arithmetics on floating-point numbers. All floating-point numbers in our benchmark cases are modeled as real numbers, disregarding precision errors.
\item \emph{List, Sequence} consists of operations involving the \texttt{list} type and Why3's \texttt{sequence}, \texttt{array31}, \texttt{array32}, and \texttt{array63}.
\item \emph{Set, Map, Bag} consists of operations whose types involve finite map, multiset, finite set, predicate-based set, and hash-table.
\item \emph{Tree, String, Matrix} consists of operations whose types involve Why3's built-in binary tree, string, and matrix.
\item \emph{Memory} consists of operations whose types involve Frama-C's memory encoding.
\item \emph{Custom Datatype} consists of operations whose types involve any datatype not provided by the system library but defined by the verification projects.
\end{itemize}
The metric \emph{depth} is the height of the abstract syntax tree of the VCs,  in the standard $\lambda$-calculus representation with all arguments of every function application represented as siblings.

\newpage

\section{Intersection Analysis of NTP and Hammer Capabilities}

\begin{wraptable}{r}{.48\textwidth}
\vspace*{-1.25em}
\centering
    \caption{The number of problems solved by both hammers and NTP models, only by hammers, and only by NTP models.}
    \label{tab:intersection}
    \small
\setlength{\tabcolsep}{4.5pt}
    \begin{tabular}{lccc}
         \toprule
         Category & Common & \parbox{1.2cm}{\centering Hammer\\only} & \parbox{0.8cm}{\centering NTP\\only} \\
         \midrule
         Algorithm & 1 & 3 & 1 \\
         Data Structure & 3 & 7 & 0 \\
         Calculation & 3 & 5 & 2 \\
         Engineering & 5 & 2 & 0 \\
         Competition & 0 & 3 & 3 \\
         Function & 6 & 19 & 0 \\
         Memory & 1 & 17 & 1 \\
         Loop & 3 & 16 & 0 \\
         Invalid Arg. & 5 & 15 & 0 \\
         \midrule
         Total & 27 & 87 & 7 \\
         \bottomrule
    \end{tabular}
\end{wraptable}

To understand whether neural and symbolic approaches overlap or diverge in their capabilities, we analyze the intersection between the union of all problems solved by NTP models and the union of all problems solved by hammers.
Table \ref{tab:intersection} presents the results.
The results reveal a complementarity: many verification conditions are solved exclusively by one method or the other.
This confirms that NTPs and hammers leverage distinct reasoning mechanisms and that neither approach is a subset of the other, highlighting the potential for hybrid solutions.

\section{Identifying VCs in CoqStop and FVEL}\label{appendix:VCnum}

In order to support the numbers given in \cref{tab:comparison}, this section describes our approach to identifying VCs in the CoqStoq benchmark~\citep{Rango} and FVEL~\citep{FVEL}.
CoqStop's test set contains 10,396 theorems from 12 Rocq projects; FVEL's test set contains 1967 cases.

\paragraph{CoqStop}

CoqStop's VCs are predominantly drawn from CompCert, which accounts for over 58\% of the test set, while other verification-related projects constitute no more than 6\%. Therefore, we focus solely on CompCert.
In CompCert, the tactics and other constructs that are relevant to program analysis and VC generation are \texttt{TransfInstr}, \texttt{UseTransfer}, \texttt{monadInv}, \texttt{step\_simulation}, \texttt{exploit}, and \texttt{match\_states}.
Among the CompCert VCs in CoqStop, only 1,325 cases involve these tactics, accounting for 12.7\% of the total test set.
Including other projects that may involve VCs (at most 6\%), the total proportion would not exceed 20\%.

\paragraph{FVEL}

All of FVEL's test cases are extracted from seL4. seL4's VCs are generated using the tactics vcg, wp, and wpsimp. Based on the test case list provided by FVEL, we analyzed cases whose proofs contain these tactics and found only 328. Therefore, the proportion of VCs in FVEL does not exceed $328/1967 < 17\%$.

%% file: main.bbl
\begin{thebibliography}{81}
\providecommand{\natexlab}[1]{#1}
\providecommand{\url}[1]{\texttt{#1}}
\expandafter\ifx\csname urlstyle\endcsname\relax
  \providecommand{\doi}[1]{doi: #1}\else
  \providecommand{\doi}{doi: \begingroup \urlstyle{rm}\Url}\fi

\bibitem[Achiam et~al.(2024)Achiam, Adler, Agarwal, Ahmad, Akkaya, Aleman, Almeida, Altenschmidt, Altman, Anadkat, et~al.]{achiam2023gpt}
Josh Achiam, Steven Adler, Sandhini Agarwal, Lama Ahmad, Ilge Akkaya, Florencia~Leoni Aleman, Diogo Almeida, Janko Altenschmidt, Sam Altman, Shyamal Anadkat, et~al.
\newblock {GPT-4 Technical Report}, 2024.
\newblock URL \url{https://arxiv.org/abs/2303.08774}.

\bibitem[Bansal et~al.(2019)Bansal, Loos, Rabe, Szegedy, and Wilcox]{HOList}
Kshitij Bansal, Sarah~M. Loos, Markus~N. Rabe, Christian Szegedy, and Stewart Wilcox.
\newblock {HOList: An Environment for Machine Learning of Higher Order Logic Theorem Proving}.
\newblock In Kamalika Chaudhuri and Ruslan Salakhutdinov (eds.), \emph{Proceedings of the 36th International Conference on Machine Learning, {ICML} 2019, 9-15 June 2019, Long Beach, California, {USA}}, volume~97 of \emph{Proceedings of Machine Learning Research}, pp.\  454--463. {PMLR}, 2019.
\newblock URL \url{http://proceedings.mlr.press/v97/bansal19a.html}.

\bibitem[Barbosa et~al.(2022)Barbosa, Barrett, Brain, Kremer, Lachnitt, Mann, Mohamed, Mohamed, Niemetz, N{\"o}tzli, Ozdemir, Preiner, Reynolds, Sheng, Tinelli, and Zohar]{CVC5}
Haniel Barbosa, Clark Barrett, Martin Brain, Gereon Kremer, Hanna Lachnitt, Makai Mann, Abdalrhman Mohamed, Mudathir Mohamed, Aina Niemetz, Andres N{\"o}tzli, Alex Ozdemir, Mathias Preiner, Andrew Reynolds, Ying Sheng, Cesare Tinelli, and Yoni Zohar.
\newblock {cvc5: A Versatile and Industrial-Strength SMT Solver}.
\newblock In Dana Fisman and Grigore Rosu (eds.), \emph{Tools and Algorithms for the Construction and Analysis of Systems}, pp.\  415--442, Cham, 2022. Springer International Publishing.
\newblock ISBN 978-3-030-99524-9.

\bibitem[Barnett et~al.(2006)Barnett, Chang, DeLine, Jacobs, and Leino]{Boogie}
Mike Barnett, Bor-Yuh~Evan Chang, Robert DeLine, Bart Jacobs, and K.~Rustan~M. Leino.
\newblock {Boogie: A Modular Reusable Verifier for Object-Oriented Programs}.
\newblock In \emph{Formal Methods for Components and Objects (FMCO)}, volume 4111 of \emph{Lecture Notes in Computer Science}, pp.\  364--387. Springer, 2006.
\newblock URL \url{https://doi.org/10.1007/11804192_17}.

\bibitem[Barrett et~al.(2010)Barrett, Stump, Tinelli, et~al.]{SMT-Lib}
Clark Barrett, Aaron Stump, Cesare Tinelli, et~al.
\newblock {The SMT-lib Standard: Version 2.0}.
\newblock In \emph{Proceedings of the 8th international workshop on satisfiability modulo theories (Edinburgh, UK)}, volume~13, pp.\ ~14, 2010.

\bibitem[Barrett et~al.(2011)Barrett, Conway, Deters, Hadarean, Jovanovi{\'{c}}, King, Reynolds, and Tinelli]{barrett2011cvc4}
Clark Barrett, Christopher~L. Conway, Morgan Deters, Liana Hadarean, Dejan Jovanovi{\'{c}}, Tim King, Andrew Reynolds, and Cesare Tinelli.
\newblock {CVC4}.
\newblock In Ganesh Gopalakrishnan and Shaz Qadeer (eds.), \emph{{Computer Aided Verification}}, pp.\  171--177, Berlin, Heidelberg, 2011. Springer Berlin Heidelberg.
\newblock ISBN 978-3-642-22110-1.

\bibitem[Barthe et~al.(2011)Barthe, Gr{\'e}goire, Heraud, and Zanella~B{\'e}guelin]{EasyCrypt}
Gilles Barthe, Benjamin Gr{\'e}goire, Sylvain Heraud, and Santiago Zanella~B{\'e}guelin.
\newblock {Computer-Aided Security Proofs for the Working Cryptographer}.
\newblock In \emph{Advances in Cryptology -- CRYPTO 2011}, volume 6841 of \emph{Lecture Notes in Computer Science}, pp.\  71--90. Springer, 2011.
\newblock URL \url{https://doi.org/10.1007/978-3-642-22792-9_5}.

\bibitem[Baudin et~al.(2021)Baudin, Bobot, B\"{u}hler, Correnson, Kirchner, Kosmatov, Maroneze, Perrelle, Prevosto, Signoles, and Williams]{FramaC}
Patrick Baudin, Fran\c{c}ois Bobot, David B\"{u}hler, Lo\"{\i}c Correnson, Florent Kirchner, Nikolai Kosmatov, Andr\'{e} Maroneze, Valentin Perrelle, Virgile Prevosto, Julien Signoles, and Nicky Williams.
\newblock {The dogged pursuit of bug-free C programs: the Frama-C software analysis platform}.
\newblock \emph{Commun. ACM}, 64\penalty0 (8):\penalty0 56–68, July 2021.
\newblock ISSN 0001-0782.
\newblock URL \url{https://doi.org/10.1145/3470569}.

\bibitem[Blanchard et~al.(2018)Blanchard, Kosmatov, and Loulergue]{ContikiList}
Allan Blanchard, Nikolai Kosmatov, and Fr{\'e}d{\'e}ric Loulergue.
\newblock {Ghosts for Lists: A Critical Module of Contiki Verified in Frama-C}.
\newblock In \emph{NASA Formal Methods (NFM 2018)}, volume 10811 of \emph{Lecture Notes in Computer Science}, pp.\  37--53. Springer, 2018.
\newblock URL \url{https://doi.org/10.1007/978-3-319-77935-5_3}.

\bibitem[Blanchette et~al.(2017)Blanchette, Meier, Popescu, and Traytel]{Isabelle-non-uniform}
Jasmin~Christian Blanchette, Fabian Meier, Andrei Popescu, and Dmitriy Traytel.
\newblock {Foundational nonuniform (Co)datatypes for higher-order logic}.
\newblock In \emph{32nd Annual {ACM/IEEE} Symposium on Logic in Computer Science, {LICS} 2017, Reykjavik, Iceland, June 20-23, 2017}, pp.\  1--12. {IEEE} Computer Society, 2017.
\newblock URL \url{https://doi.org/10.1109/LICS.2017.8005071}.

\bibitem[Bobot et~al.(2025)Bobot, Filli\^atre, March\'e, Melquiond, and Paskevich]{Why3Manual}
Fran\c{c}ois Bobot, Jean-Christophe Filli\^atre, Claude March\'e, Guillaume Melquiond, and Andrei Paskevich.
\newblock \emph{{The Why3 Platform}}.
\newblock University Paris\textendash Saclay, CNRS, Inria, version 1.8.2 edition, September 2025.
\newblock URL \url{https://www.why3.org/doc/}.

\bibitem[Bobot et~al.()Bobot, Filliâtre, Marché, Melquiond, and Paskevich]{AL3}
François Bobot, Jean-Christophe Filliâtre, Claude Marché, Guillaume Melquiond, and Andrei Paskevich.
\newblock {Basic Strategies}.
\newblock \url{https://www.why3.org/doc/technical.html#basic-strategies}.
\newblock Accessed: 2026-01-26.

\bibitem[B{\"o}hme \& Nipkow(2010)B{\"o}hme and Nipkow]{SLEDGEHAMMER}
Sascha B{\"o}hme and Tobias Nipkow.
\newblock {Sledgehammer: Judgement Day}.
\newblock In \emph{Automated Reasoning (IJCAR 2010)}, volume 6173 of \emph{Lecture Notes in Computer Science}, pp.\  107--121. Springer, 2010.
\newblock URL \url{https://doi.org/10.1007/978-3-642-14203-1_9}.

\bibitem[Burghardt et~al.(2015)Burghardt, Gerlach, and Lapawczyk]{burghardt2015acsl}
Jochen Burghardt, Jens Gerlach, and Timon Lapawczyk.
\newblock {ACSL} by example.
\newblock Technical report, Fraunhofer FOKUS, 2015.
\newblock URL \url{https://publica.fraunhofer.de/entities/publication/beb926ba-c3d6-4570-acc6-dd50da41843f}.

\bibitem[Carvalho et~al.(2014)Carvalho, da~Silva~Sousa, Pinto, and Tomb]{kLIBCv}
Nuno Carvalho, Cristiano da~Silva~Sousa, Jorge~Sousa Pinto, and Aaron Tomb.
\newblock {Formal Verification of kLIBC with the WP Frama-C Plug-in}.
\newblock In Julia~M. Badger and Kristin~Yvonne Rozier (eds.), \emph{NASA Formal Methods (NFM 2014)}, volume 8430 of \emph{Lecture Notes in Computer Science}, pp.\  343--358. Springer, 2014.
\newblock \doi{https://doi.org/10.1007/978-3-319-06200-6_29}.

\bibitem[Cheng et~al.(2025)Cheng, Fan, Hao, Killian, Li, Sun, Ren, Moreno, Zhang, Zhong, Xiong, Hu, Xie, Han, Wang, Pimpalkhute, Zhuang, Singh, Liang, Xie, She, Fan, Gao, Ma, Yurochkin, Maggs, Ma, He, Hu, Liu, and Xing]{cheng2025k2}
Zhoujun Cheng, Richard Fan, Shibo Hao, Taylor~W. Killian, Haonan Li, Suqi Sun, Hector Ren, Alexander Moreno, Daqian Zhang, Tianjun Zhong, Yuxin Xiong, Yuanzhe Hu, Yutao Xie, Xudong Han, Yuqi Wang, Varad Pimpalkhute, Yonghao Zhuang, Aaryamonvikram Singh, Xuezhi Liang, Anze Xie, Jianshu She, Desai Fan, Chengqian Gao, Liqun Ma, Mikhail Yurochkin, John Maggs, Xuezhe Ma, Guowei He, Zhiting Hu, Zhengzhong Liu, and Eric~P. Xing.
\newblock K2-think: A parameter-efficient reasoning system, 2025.
\newblock URL \url{https://arxiv.org/abs/2509.07604}.

\bibitem[Church(1940)]{Church_1940}
Alonzo Church.
\newblock A formulation of the simple theory of types.
\newblock \emph{Journal of Symbolic Logic}, 5\penalty0 (2):\penalty0 56–68, 1940.
\newblock URL \url{https://doi.org/10.2307/2266170}.

\bibitem[Cohen et~al.(2009)Cohen, Dahlweid, Hillebrand, Leinenbach, Moskal, Santen, Schulte, and Tobies]{VCC}
Ernie Cohen, Markus Dahlweid, Mark Hillebrand, Dirk Leinenbach, Micha{\l} Moskal, Thomas Santen, Wolfram Schulte, and Stephan Tobies.
\newblock {VCC}: A practical system for verifying concurrent {C}.
\newblock In \emph{Theorem Proving in Higher Order Logics (TPHOLs)}, volume 5674 of \emph{Lecture Notes in Computer Science}, pp.\  23--42. Springer, 2009.
\newblock URL \url{https://doi.org/10.1007/978-3-642-03359-9_2}.

\bibitem[Conchon et~al.(2018)Conchon, Coquereau, Iguernlala, and Mebsout]{AltErgo}
Sylvain Conchon, Albin Coquereau, Mohamed Iguernlala, and Alain Mebsout.
\newblock {Alt-Ergo 2.2}.
\newblock In \emph{{SMT Workshop: International Workshop on Satisfiability Modulo Theories}}, Oxford, United Kingdom, July 2018.
\newblock URL \url{https://inria.hal.science/hal-01960203}.

\bibitem[Coquand \& Huet(1988)Coquand and Huet]{CoquandHuet1988CoC}
Thierry Coquand and G{\'e}rard Huet.
\newblock {The Calculus of Constructions}.
\newblock \emph{Information and Computation}, 76\penalty0 (2--3):\penalty0 95--120, 1988.
\newblock URL \url{https://doi.org/10.1016/0890-5401(88)90005-3}.

\bibitem[Correnson et~al.(2025)Correnson, Cuoq, Kirchner, Maroneze, Prevosto, Puccetti, Signoles, and Yakobowski]{FramaCManual}
Lo\"ic Correnson, Pascal Cuoq, Florent Kirchner, Andr\'e Maroneze, Virgile Prevosto, Armand Puccetti, Julien Signoles, and Boris Yakobowski.
\newblock \emph{{Frama-C User Manual}}, 2025.
\newblock URL \url{https://frama-c.com/download/frama-c-user-manual.pdf}.
\newblock Corresponds to Frama-C 31.0 (Gallium), released on 2025-06-24.

\bibitem[Cousot et~al.(2005)Cousot, Cousot, Feret, Mauborgne, Min{\'e}, Monniaux, and Rival]{cousot2005astree}
Patrick Cousot, Radhia Cousot, Jer{\^o}me Feret, Laurent Mauborgne, Antoine Min{\'e}, David Monniaux, and Xavier Rival.
\newblock {The ASTRE\'E Analyzer}.
\newblock In Mooly Sagiv (ed.), \emph{Programming Languages and Systems}, pp.\  21--30, Berlin, Heidelberg, 2005. Springer Berlin Heidelberg.
\newblock ISBN 978-3-540-31987-0.
\newblock \doi{https://doi.org/10.1007/978-3-540-31987-0_3}.

\bibitem[Czajka \& Kaliszyk(2018)Czajka and Kaliszyk]{COQHAMMER}
{\L}ukasz Czajka and Cezary Kaliszyk.
\newblock {Hammer for Coq: Automation for Dependent Type Theory}.
\newblock \emph{Journal of Automated Reasoning}, 61\penalty0 (1--4):\penalty0 423--453, 2018.
\newblock URL \url{https://doi.org/10.1007/s10817-018-9458-4}.

\bibitem[de~Moura \& Ullrich(2021)de~Moura and Ullrich]{Lean4}
Leonardo de~Moura and Sebastian Ullrich.
\newblock {The Lean 4 Theorem Prover and Programming Language}.
\newblock In \emph{Automated Deduction -- CADE 28}, volume 12699 of \emph{Lecture Notes in Computer Science}, pp.\  625--635. Springer, 2021.
\newblock URL \url{https://doi.org/10.1007/978-3-030-79876-5_37}.

\bibitem[de~Moura \& Bj{\o}rner(2008)de~Moura and Bj{\o}rner]{Z3}
Leonardo~Mendon{\c{c}}a de~Moura and Nikolaj~S. Bj{\o}rner.
\newblock {Z3:} an efficient {SMT} solver.
\newblock In C.~R. Ramakrishnan and Jakob Rehof (eds.), \emph{Tools and Algorithms for the Construction and Analysis of Systems, 14th International Conference, {TACAS} 2008, Held as Part of the Joint European Conferences on Theory and Practice of Software, {ETAPS} 2008, Budapest, Hungary, March 29-April 6, 2008. Proceedings}, volume 4963 of \emph{Lecture Notes in Computer Science}, pp.\  337--340. Springer, 2008.
\newblock URL \url{https://doi.org/10.1007/978-3-540-78800-3_24}.

\bibitem[DeepSeek-AI et~al.(2025)DeepSeek-AI, Liu, Feng, Xue, Wang, Wu, Lu, Zhao, Deng, Zhang, Ruan, Dai, Guo, Yang, Chen, Ji, Li, et~al.]{liu2024deepseek}
DeepSeek-AI, Aixin Liu, Bei Feng, Bing Xue, Bingxuan Wang, Bochao Wu, Chengda Lu, Chenggang Zhao, Chengqi Deng, Chenyu Zhang, Chong Ruan, Damai Dai, Daya Guo, Dejian Yang, Deli Chen, Dongjie Ji, Erhang Li, et~al.
\newblock Deepseek-v3 technical report, 2025.
\newblock URL \url{https://arxiv.org/abs/2412.19437}.

\bibitem[Denis et~al.(2022)Denis, Jourdan, and March{\'e}]{Creusot}
Xavier Denis, Jacques-Henri Jourdan, and Claude March{\'e}.
\newblock {Creusot: A Foundry for the Deductive Verification of Rust Programs}.
\newblock In \emph{Formal Methods and Software Engineering (ICFEM)}, volume 13478 of \emph{Lecture Notes in Computer Science}, pp.\  90--105. Springer, 2022.
\newblock URL \url{https://doi.org/10.1007/978-3-031-17244-1_6}.

\bibitem[Djoudi et~al.(2021)Djoudi, H{\'a}na, and Kosmatov]{JavaCard}
Adel Djoudi, Martin H{\'a}na, and Nikolai Kosmatov.
\newblock {Formal Verification of a JavaCard Virtual Machine with Frama-C}.
\newblock In \emph{Formal Methods (FM 2021)}, volume 13047 of \emph{Lecture Notes in Computer Science}, pp.\  427--444. Springer, 2021.
\newblock URL \url{https://doi.org/10.1007/978-3-030-90870-6_23}.

\bibitem[Dougherty \& Mehta(2025)Dougherty and Mehta]{DoughertyMehta2025}
Quinn Dougherty and Ronak Mehta.
\newblock Proving the coding interview: A benchmark for formally verified code generation.
\newblock In \emph{2025 IEEE/ACM 2nd International Workshop on Large Language Models for Code (LLM4Code)}, pp.\  72--79. IEEE, 2025.
\newblock URL \url{https://doi.org/10.1109/LLM4Code66737.2025.00014}.

\bibitem[Dutle et~al.(2021)Dutle, Moscato, Titolo, Mu{\~{n}}oz, Anderson, and Bobot]{Dutle2021}
Aaron Dutle, Mariano Moscato, Laura Titolo, C{\'e}sar Mu{\~{n}}oz, Gregory Anderson, and Fran{\c{c}}ois Bobot.
\newblock Formal analysis of the compact positionreporting algorithm.
\newblock \emph{Formal Aspects of Computing}, 33\penalty0 (1):\penalty0 65--86, Jan 2021.
\newblock ISSN 1433-299X.
\newblock URL \url{https://doi.org/10.1007/s00165-019-00504-0}.

\bibitem[Ebalard et~al.(2019)Ebalard, Mouy, and Benadjila]{EbalardMouyBenadjila}
Arnaud Ebalard, Patricia Mouy, and Ryad Benadjila.
\newblock {Journey to a RTE-free X.509 parser}.
\newblock In \emph{Symposium sur la s{\'e}curit{\'e} des technologies de l'information et des communications (SSTIC 2019)}, pp.\  1--50, Rennes, France, 2019.
\newblock URL \url{https://www.sstic.org/media/SSTIC2019/SSTIC-actes/journey-to-a-rte-free-x509-parser/SSTIC2019-Article-journey-to-a-rte-free-x509-parser-ebalard_mouy_benadjila_3cUxSCv.pdf}.
\newblock Talk on 6 June 2019; paper PDF available via the presentation page.

\bibitem[Efremov et~al.(2018)Efremov, Mandrykin, and Khoroshilov]{verker}
Denis Efremov, Mikhail Mandrykin, and Alexey Khoroshilov.
\newblock {Deductive Verification of Unmodified {Linux} Kernel Library Functions}.
\newblock In Tiziana Margaria and Bernhard Steffen (eds.), \emph{Leveraging Applications of Formal Methods, Verification and Validation. Verification (ISoLA 2018)}, volume 11245 of \emph{Lecture Notes in Computer Science}, pp.\  216--234. Springer, 2018.
\newblock URL \url{https://doi.org/10.1007/978-3-030-03421-4_15}.

\bibitem[Ernst et~al.(2019)Ernst, Huisman, Mostowski, and Ulbrich]{VerifyThis}
Gidon Ernst, Marieke Huisman, Wojciech Mostowski, and Mattias Ulbrich.
\newblock {VerifyThis - Verification Competition with a Human Factor}.
\newblock In Dirk Beyer, Marieke Huisman, Fabrice Kordon, and Bernhard Steffen (eds.), \emph{Tools and Algorithms for the Construction and Analysis of Systems - 25 Years of {TACAS:} TOOLympics, Held as Part of {ETAPS} 2019, Prague, Czech Republic, April 6-11, 2019, Proceedings, Part {III}}, volume 11429 of \emph{Lecture Notes in Computer Science}, pp.\  176--195. Springer, 2019.
\newblock URL \url{https://doi.org/10.1007/978-3-030-17502-3_12}.

\bibitem[Filli\^atre \& Paskevich(2013)Filli\^atre and Paskevich]{Why3}
Jean-Christophe Filli\^atre and Andrei Paskevich.
\newblock {Why3 --- Where Programs Meet Provers}.
\newblock In \emph{Programming Languages and Systems (ESOP 2013)}, volume 7792 of \emph{Lecture Notes in Computer Science}, pp.\  125--128. Springer, 2013.
\newblock URL \url{https://doi.org/10.1007/978-3-642-37036-6_8}.

\bibitem[Garavel et~al.(2020)Garavel, ter Beek, and van~de Pol]{ExpertSurvey}
Hubert Garavel, Maurice~H. ter Beek, and Jaco van~de Pol.
\newblock {The 2020 Expert Survey on Formal Methods}.
\newblock In \emph{Formal Methods for Industrial Critical Systems (FMICS)}, volume 12327 of \emph{Lecture Notes in Computer Science}, pp.\  3--69. Springer, 2020.
\newblock URL \url{https://doi.org/10.1007/978-3-030-58298-2_1}.

\bibitem[Gauthier et~al.(2021)Gauthier, Kaliszyk, Urban, Kumar, and Norrish]{tactictoe}
Thibault Gauthier, Cezary Kaliszyk, Josef Urban, Ramana Kumar, and Michael Norrish.
\newblock {TacticToe: Learning to Prove with Tactics}.
\newblock \emph{J. Autom. Reason.}, 65\penalty0 (2):\penalty0 257–286, February 2021.
\newblock ISSN 0168-7433.
\newblock URL \url{https://doi.org/10.1007/s10817-020-09580-x}.

\bibitem[Harrison et~al.(2014)Harrison, Urban, and Wiedijk]{HARRISON2014135}
John Harrison, Josef Urban, and Freek Wiedijk.
\newblock {History of Interactive Theorem Proving}.
\newblock In J{\"o}rg~H. Siekmann (ed.), \emph{Computational Logic}, volume~9 of \emph{Handbook of the History of Logic}, pp.\  135--214. North-Holland, 2014.
\newblock URL \url{https://doi.org/10.1016/B978-0-444-51624-4.50004-6}.

\bibitem[Hoare(1969)]{HoareTriple}
C.~A.~R. Hoare.
\newblock An axiomatic basis for computer programming.
\newblock \emph{Commun. ACM}, 12\penalty0 (10):\penalty0 576–580, October 1969.
\newblock ISSN 0001-0782.
\newblock URL \url{https://doi.org/10.1145/363235.363259}.

\bibitem[Huang et~al.(2019)Huang, Dhariwal, Song, and Sutskever]{GamePad}
Daniel Huang, Prafulla Dhariwal, Dawn Song, and Ilya Sutskever.
\newblock {GamePad: A Learning Environment for Theorem Proving}.
\newblock In \emph{7th International Conference on Learning Representations, {ICLR} 2019, New Orleans, LA, USA, May 6-9, 2019}. OpenReview.net, 2019.
\newblock URL \url{https://openreview.net/forum?id=r1xwKoR9Y7}.

\bibitem[Kaliszyk et~al.(2017)Kaliszyk, Chollet, and Szegedy]{HOLStep}
Cezary Kaliszyk, Fran{\c{c}}ois Chollet, and Christian Szegedy.
\newblock {HolStep: A Machine Learning Dataset for Higher-order Logic Theorem Proving}.
\newblock In \emph{5th International Conference on Learning Representations, {ICLR} 2017, Toulon, France, April 24-26, 2017, Conference Track Proceedings}. OpenReview.net, 2017.
\newblock URL \url{https://openreview.net/forum?id=ryuxYmvel}.

\bibitem[Klebanov et~al.(2011)Klebanov, M{\"{u}}ller, Shankar, Leavens, W{\"{u}}stholz, Alkassar, Arthan, Bronish, Chapman, Cohen, Hillebrand, Jacobs, Leino, Monahan, Piessens, Polikarpova, Ridge, Smans, Tobies, Tuerk, Ulbrich, and Wei{\ss}]{VSCOMP}
Vladimir Klebanov, Peter M{\"{u}}ller, Natarajan Shankar, Gary~T. Leavens, Valentin W{\"{u}}stholz, Eyad Alkassar, Rob Arthan, Derek Bronish, Rod Chapman, Ernie Cohen, Mark~A. Hillebrand, Bart Jacobs, K.~Rustan~M. Leino, Rosemary Monahan, Frank Piessens, Nadia Polikarpova, Tom Ridge, Jan Smans, Stephan Tobies, Thomas Tuerk, Mattias Ulbrich, and Benjamin Wei{\ss}.
\newblock {The 1st Verified Software Competition: Experience Report}.
\newblock In Michael~J. Butler and Wolfram Schulte (eds.), \emph{{FM} 2011: Formal Methods - 17th International Symposium on Formal Methods, Limerick, Ireland, June 20-24, 2011. Proceedings}, volume 6664 of \emph{Lecture Notes in Computer Science}, pp.\  154--168. Springer, 2011.
\newblock URL \url{https://doi.org/10.1007/978-3-642-21437-0_14}.

\bibitem[Kov{\'a}cs \& Voronkov(2013)Kov{\'a}cs and Voronkov]{kovacs2013first}
Laura Kov{\'a}cs and Andrei Voronkov.
\newblock {First-Order Theorem Proving and Vampire}.
\newblock In Natasha Sharygina and Helmut Veith (eds.), \emph{{Computer Aided Verification}}, pp.\  1--35, Berlin, Heidelberg, 2013. Springer Berlin Heidelberg.
\newblock ISBN 978-3-642-39799-8.

\bibitem[Lattuada et~al.(2023)Lattuada, Hance, Cho, Brun, Subasinghe, Zhou, Howell, Parno, and Hawblitzel]{Verus}
Andrea Lattuada, Travis Hance, Chanhee Cho, Matthias Brun, Isitha Subasinghe, Yi~Zhou, Jon Howell, Bryan Parno, and Chris Hawblitzel.
\newblock {Verus: Verifying Rust Programs using Linear Ghost Types}.
\newblock \emph{Proc. ACM Program. Lang.}, 7\penalty0 (OOPSLA1), April 2023.
\newblock URL \url{https://doi.org/10.1145/3586037}.

\bibitem[Lawall et~al.(2025)Lawall, Nishimura, and Lozi]{ShouldWeBalance}
Julia Lawall, Keisuke Nishimura, and Jean-Pierre Lozi.
\newblock {Should We Balance? Towards Formal Verification of the Linux Kernel Scheduler}.
\newblock In Roberto Giacobazzi and Alessandra Gorla (eds.), \emph{Static Analysis (SAS 2024)}, volume 14995 of \emph{Lecture Notes in Computer Science}, pp.\  194--215. Springer, 2025.
\newblock URL \url{https://doi.org/10.1007/978-3-031-74776-2_8}.

\bibitem[Leino(2010)]{Dafny}
K.~Rustan~M. Leino.
\newblock {Dafny: An Automatic Program Verifier for Functional Correctness}.
\newblock In \emph{Logic for Programming, Artificial Intelligence, and Reasoning (LPAR)}, volume 6355 of \emph{Lecture Notes in Computer Science}, pp.\  348--370. Springer, 2010.
\newblock URL \url{https://doi.org/10.1007/978-3-642-17511-4_20}.

\bibitem[Li et~al.(2021)Li, Yu, Wu, and Paulson]{LiYuWuPaulson2021IsarStep}
Wenda Li, Lei Yu, Yuhuai Wu, and Lawrence~C. Paulson.
\newblock {IsarStep: a Benchmark for High-level Mathematical Reasoning}.
\newblock In \emph{International Conference on Learning Representations}, 2021.
\newblock URL \url{https://openreview.net/forum?id=Pzj6fzU6wkj}.

\bibitem[Lin et~al.(2024)Lin, Cao, Huang, Wang, Lu, Liu, Song, and Liang]{FVEL}
Xiaohan Lin, Qingxing Cao, Yinya Huang, Haiming Wang, Jianqiao Lu, Zhengying Liu, Linqi Song, and Xiaodan Liang.
\newblock {FVEL}: Interactive formal verification environment with large language models via theorem proving.
\newblock In A.~Globerson, L.~Mackey, D.~Belgrave, A.~Fan, U.~Paquet, J.~Tomczak, and C.~Zhang (eds.), \emph{Advances in Neural Information Processing Systems}, volume~37, pp.\  54932--54946. Curran Associates, Inc., 2024.
\newblock URL \url{https://doi.org/10.52202/079017-1742}.

\bibitem[Lin et~al.(2025)Lin, Tang, Lyu, Wu, Lin, Yang, Li, Xia, Chen, Arora, and Jin]{lin2025goedel}
Yong Lin, Shange Tang, Bohan Lyu, Jiayun Wu, Hongzhou Lin, Kaiyu Yang, Jia Li, Mengzhou Xia, Danqi Chen, Sanjeev Arora, and Chi Jin.
\newblock {Goedel-Prover: A Frontier Model for Open-Source Automated Theorem Proving}, 2025.
\newblock URL \url{https://arxiv.org/abs/2502.07640}.

\bibitem[Lohn \& Welleck(2024)Lohn and Welleck]{miniCodeProps}
Evan Lohn and Sean Welleck.
\newblock {miniCodeProps: a Minimal Benchmark for Proving Code Properties}.
\newblock In \emph{Neurips Safe Generative AI Workshop 2024}, 2024.
\newblock URL \url{https://openreview.net/forum?id=6QFe3vPbYZ}.

\bibitem[Loughridge et~al.(2025)Loughridge, Sun, Ahrenbach, Cassano, Sun, Sheng, Mudide, Misu, Amin, and Tegmark]{DafnyBench}
Chloe~R Loughridge, Qinyi Sun, Seth Ahrenbach, Federico Cassano, Chuyue Sun, Ying Sheng, Anish Mudide, Md~Rakib~Hossain Misu, Nada Amin, and Max Tegmark.
\newblock {DafnyBench: A Benchmark for Formal Software Verification}.
\newblock \emph{Transactions on Machine Learning Research}, 2025.
\newblock ISSN 2835-8856.
\newblock URL \url{https://openreview.net/forum?id=yBgTVWccIx}.

\bibitem[Mangano et~al.(2017)Mangano, Duquennoy, and Kosmatov]{ContikiMem}
Fr{\'e}d{\'e}ric Mangano, Simon Duquennoy, and Nikolai Kosmatov.
\newblock {Formal Verification of a Memory Allocation Module of Contiki with Frama-C: A Case Study}.
\newblock In Fr{\'e}d{\'e}ric Cuppens, Nora Cuppens, Jean-Louis Lanet, and Axel Legay (eds.), \emph{Risks and Security of Internet and Systems}, pp.\  114--120, Cham, 2017. Springer International Publishing.
\newblock ISBN 978-3-319-54876-0.

\bibitem[Minervini et~al.(2018)Minervini, Bosnjak, Rocktäschel, and Riedel]{minervini2018towards}
Pasquale Minervini, Matko Bosnjak, Tim Rocktäschel, and Sebastian Riedel.
\newblock {Towards Neural Theorem Proving at Scale}.
\newblock \emph{CoRR}, abs/1807.08204, 2018.
\newblock URL \url{http://arxiv.org/abs/1807.08204}.

\bibitem[Mugnier et~al.(2025)Mugnier, Gonzalez, Polikarpova, Jhala, and Zhou]{Laurel}
Eric Mugnier, Emmanuel~Anaya Gonzalez, Nadia Polikarpova, Ranjit Jhala, and Yuanyuan Zhou.
\newblock {Laurel: Unblocking Automated Verification with Large Language Models}.
\newblock \emph{Proc. ACM Program. Lang.}, 9\penalty0 (OOPSLA1), April 2025.
\newblock URL \url{https://doi.org/10.1145/3720499}.

\bibitem[Paulson(1990)]{Isabelle}
Lawrence~C. Paulson.
\newblock {Isabelle: The Next 700 Theorem Provers}.
\newblock In Piergiorgio Odifreddi (ed.), \emph{Logic and Computer Science}, pp.\  361--386. Academic Press, London, 1990.

\bibitem[Pereira \& Ravara(2021)Pereira and Ravara]{Cameleer}
M{\'a}rio Pereira and Ant{\'o}nio Ravara.
\newblock {Cameleer: A Deductive Verification Tool for OCaml}.
\newblock In \emph{Computer Aided Verification (CAV)}, volume 12760 of \emph{Lecture Notes in Computer Science}, pp.\  677--689. Springer, 2021.
\newblock URL \url{https://doi.org/10.1007/978-3-030-81688-9_31}.

\bibitem[Peyrard et~al.(2018)Peyrard, Kosmatov, Duquennoy, Lille, and Raza]{AESCCM}
Alexandre Peyrard, Nikolai Kosmatov, Simon Duquennoy, Inria Lille, and Shahid Raza.
\newblock {Towards Formal Verification of Contiki: Analysis of the AES-CCM* Modules with Frama-C}.
\newblock In \emph{Proceedings of the 2018 International Conference on Embedded Wireless Systems and Networks}, EWSN ’18, pp.\  264–269. International Conference on Embedded Wireless Systems and Networks (EWSN), 2018.
\newblock ISBN 9780994988621.

\bibitem[Pollien et~al.(2021)Pollien, Garion, Hattenberger, Roux, and Thirioux]{paparazzi}
Baptiste Pollien, Christophe Garion, Gautier Hattenberger, Pierre Roux, and Xavier Thirioux.
\newblock {Verifying the Mathematical Library of an UAV Autopilot with Frama-C}.
\newblock In Alberto Lluch~Lafuente and Anastasia Mavridou (eds.), \emph{Formal Methods for Industrial Critical Systems (FMICS 2021)}, volume 12863 of \emph{Lecture Notes in Computer Science}, pp.\  167--173. Springer, 2021.
\newblock URL \url{https://doi.org/10.1007/978-3-030-85248-1_10}.

\bibitem[Ren et~al.(2025)Ren, Shao, Song, Xin, Wang, Zhao, Zhang, Fu, Zhu, Yang, Wu, Gou, Ma, Tang, Liu, Gao, Guo, and Ruan]{ren2025deepseek}
Z.~Z. Ren, Zhihong Shao, Junxiao Song, Huajian Xin, Haocheng Wang, Wanjia Zhao, Liyue Zhang, Zhe Fu, Qihao Zhu, Dejian Yang, Z.~F. Wu, Zhibin Gou, Shirong Ma, Hongxuan Tang, Yuxuan Liu, Wenjun Gao, Daya Guo, and Chong Ruan.
\newblock {DeepSeek-Prover-V2: Advancing Formal Mathematical Reasoning via Reinforcement Learning for Subgoal Decomposition}, 2025.
\newblock URL \url{https://arxiv.org/abs/2504.21801}.

\bibitem[Rushby(2009)]{rushby2009software}
John Rushby.
\newblock {Software Verification and System Assurance}.
\newblock In \emph{2009 Seventh IEEE International Conference on Software Engineering and Formal Methods}, pp.\  3--10, 2009.
\newblock URL \url{https://doi.org/10.1109/SEFM.2009.39}.

\bibitem[Schulz(2002)]{Eprover}
Stephan Schulz.
\newblock {E -- A Brainiac Theorem Prover}.
\newblock \emph{AI Communications}, 15\penalty0 (2-3):\penalty0 111--126, 2002.
\newblock URL \url{https://doi.org/10.3233/EAI-2002-260}.

\bibitem[Schurr et~al.(2021)Schurr, Fleury, and Desharnais]{schurr2021reliable}
Hans-J{\"o}rg Schurr, Mathias Fleury, and Martin Desharnais.
\newblock {Reliable Reconstruction of Fine-grained Proofs in a Proof Assistant}.
\newblock In Andr{\'e} Platzer and Geoff Sutcliffe (eds.), \emph{Automated Deduction -- CADE 28}, pp.\  450--467, Cham, 2021. Springer International Publishing.
\newblock ISBN 978-3-030-79876-5.
\newblock URL \url{https://doi.org/10.1007/978-3-030-79876-5_26}.

\bibitem[{SMT-LIB Initiative}()]{smtlib_logics}
{SMT-LIB Initiative}.
\newblock {SMT-LIB} --- logics.
\newblock \url{https://smt-lib.org/logics.shtml}.
\newblock Accessed: 2025-11-18.

\bibitem[Sun et~al.(2024)Sun, Sheng, Padon, and Barrett]{Clover}
Chuyue Sun, Ying Sheng, Oded Padon, and Clark~W. Barrett.
\newblock {Clover: Closed-Loop Verifiable Code Generation}.
\newblock In Guy Avni, Mirco Giacobbe, Taylor~T. Johnson, Guy Katz, Anna Lukina, Nina Narodytska, and Christian Schilling (eds.), \emph{{AI} Verification - First International Symposium, {SAIV} 2024, Montreal, QC, Canada, July 22-23, 2024, Proceedings}, volume 14846 of \emph{Lecture Notes in Computer Science}, pp.\  134--155. Springer, 2024.
\newblock URL \url{https://doi.org/10.1007/978-3-031-65112-0\_7}.

\bibitem[Sutcliffe(2024)]{TPTP}
Geoff Sutcliffe.
\newblock {Stepping Stones in the TPTP World}.
\newblock In Christoph Benzm{\"u}ller, Marijn~J.H. Heule, and Renate~A. Schmidt (eds.), \emph{Automated Reasoning}, volume 14739, pp.\  30--50. Springer Nature Switzerland, 2024.
\newblock ISBN 978-3-031-63498-7.
\newblock \doi{10.1007/978-3-031-63498-7_3}.

\bibitem[Thakur et~al.(2025)Thakur, Lee, Tsoukalas, Sistla, Zhao, Zetzsche, Durrett, Yue, and Chaudhuri]{CLEVER}
Amitayush Thakur, Jasper Lee, George Tsoukalas, Meghana Sistla, Matthew Zhao, Stefan Zetzsche, Greg Durrett, Yisong Yue, and Swarat Chaudhuri.
\newblock {CLEVER: A Curated Benchmark for Formally Verified Code Generation}, 2025.
\newblock URL \url{https://arxiv.org/abs/2505.13938}.

\bibitem[Thompson et~al.(2025)Thompson, Saavedra, Carrott, Fisher, Sanchez{-}Stern, Brun, Ferreira, Lerner, and First]{Rango}
Kyle Thompson, Nuno Saavedra, Pedro Carrott, Kevin Fisher, Alex Sanchez{-}Stern, Yuriy Brun, Jo{\~{a}}o~F. Ferreira, Sorin Lerner, and Emily First.
\newblock {Rango: Adaptive Retrieval-Augmented Proving for Automated Software Verification}.
\newblock In \emph{47th {IEEE/ACM} International Conference on Software Engineering, {ICSE} 2025, Ottawa, ON, Canada, April 26 - May 6, 2025}, pp.\  347--359. {IEEE}, 2025.
\newblock URL \url{https://doi.org/10.1109/ICSE55347.2025.00161}.

\bibitem[{Toccata Team}(2025)]{toccata_gallery}
{Toccata Team}.
\newblock Gallery of verified programs.
\newblock \url{https://toccata.gitlabpages.inria.fr/toccata/gallery/index.en.html}, 2025.
\newblock Joint team of Inria, CNRS and University of Paris-Saclay. Page generated on 2025-09-16. Accessed: 2025-09-17.

\bibitem[Tsoukalas et~al.(2024)Tsoukalas, Lee, Jennings, Xin, Ding, Jennings, Thakur, and Chaudhuri]{PutnamBench}
George Tsoukalas, Jasper Lee, John Jennings, Jimmy Xin, Michelle Ding, Michael Jennings, Amitayush Thakur, and Swarat Chaudhuri.
\newblock {PutnamBench: Evaluating Neural Theorem-Provers on the Putnam Mathematical Competition}.
\newblock In A.~Globerson, L.~Mackey, D.~Belgrave, A.~Fan, U.~Paquet, J.~Tomczak, and C.~Zhang (eds.), \emph{Advances in Neural Information Processing Systems}, volume~37, pp.\  11545--11569. Curran Associates, Inc., 2024.
\newblock \doi{10.52202/079017-0368}.
\newblock URL \url{https://proceedings.neurips.cc/paper_files/paper/2024/file/1582eaf9e0cf349e1e5a6ee453100aa1-Paper-Datasets_and_Benchmarks_Track.pdf}.

\bibitem[Vukmirovi{\'{c}} et~al.(2022)Vukmirovi{\'{c}}, Bentkamp, Blanchette, Cruanes, Nummelin, and Tourret]{vukmirovic2021making}
Petar Vukmirovi{\'{c}}, Alexander Bentkamp, Jasmin Blanchette, Simon Cruanes, Visa Nummelin, and Sophie Tourret.
\newblock {Making Higher-Order Superposition Work}.
\newblock \emph{Journal of Automated Reasoning}, 66\penalty0 (4):\penalty0 541--564, Nov 2022.
\newblock ISSN 1573-0670.
\newblock URL \url{https://doi.org/10.1007/s10817-021-09613-z}.

\bibitem[Wayne(2018)]{Hillel}
Hillel Wayne.
\newblock {The Great Theorem Prover Showdown}, 2018.
\newblock URL \url{https://www.hillelwayne.com/post/theorem-prover-showdown/}.
\newblock Blog post. Accessed: 2025-09-14.

\bibitem[Weidenbach et~al.(2009)Weidenbach, Dimova, Fietzke, Kumar, Suda, and Wischnewski]{SPASS}
Christoph Weidenbach, Dilyana Dimova, Arnaud Fietzke, Rohit Kumar, Martin Suda, and Patrick Wischnewski.
\newblock {SPASS} version 3.5.
\newblock In Renate~A. Schmidt (ed.), \emph{Automated Deduction - CADE-22, 22nd International Conference on Automated Deduction, Montreal, Canada, August 2-7, 2009. Proceedings}, volume 5663 of \emph{Lecture Notes in Computer Science}, pp.\  140--145. Springer, 2009.
\newblock URL \url{https://doi.org/10.1007/978-3-642-02959-2_10}.

\bibitem[Woodcock et~al.(2009)Woodcock, Larsen, Bicarregui, and Fitzgerald]{woodcock2009formal}
Jim Woodcock, Peter~Gorm Larsen, Juan Bicarregui, and John Fitzgerald.
\newblock {Formal methods: Practice and experience}.
\newblock \emph{ACM Comput. Surv.}, 41\penalty0 (4), October 2009.
\newblock ISSN 0360-0300.
\newblock URL \url{https://doi.org/10.1145/1592434.1592436}.

\bibitem[Xin et~al.(2025)Xin, Li, Jin, Fleuriot, and Li]{APE}
Huajian Xin, Luming Li, Xiaoran Jin, Jacques Fleuriot, and Wenda Li.
\newblock {APE-Bench I: Towards File-level Automated Proof Engineering of Formal Math Libraries}, 2025.
\newblock URL \url{https://arxiv.org/abs/2504.19110}.

\bibitem[Xu et~al.(2025)Xu, Wang, Wang, Li, and Watt]{Minilang}
Qiyuan Xu, Renxi Wang, Peixin Wang, Haonan Li, and Conrad Watt.
\newblock {A Minimalist Proof Language for Neural Theorem Proving over Isabelle/HOL}, 2025.
\newblock URL \url{https://arxiv.org/abs/2507.18885}.

\bibitem[Yang et~al.(2025)Yang, Li, Yang, Zhang, Hui, Zheng, Yu, Gao, Huang, Lv, Zheng, Liu, Zhou, Huang, Hu, Ge, Wei, Lin, Tang, Yang, Tu, Zhang, Yang, Yang, Zhou, Zhou, Lin, Dang, Bao, Yang, Yu, Deng, Li, Xue, Li, Zhang, Wang, Zhu, Men, Gao, Liu, Luo, Li, Tang, Yin, Ren, Wang, Zhang, Ren, Fan, Su, Zhang, Zhang, Wan, Liu, Wang, Cui, Zhang, Zhou, and Qiu]{yang2025qwen3}
An~Yang, Anfeng Li, Baosong Yang, Beichen Zhang, Binyuan Hui, Bo~Zheng, Bowen Yu, Chang Gao, Chengen Huang, Chenxu Lv, Chujie Zheng, Dayiheng Liu, Fan Zhou, Fei Huang, Feng Hu, Hao Ge, Haoran Wei, Huan Lin, Jialong Tang, Jian Yang, Jianhong Tu, Jianwei Zhang, Jianxin Yang, Jiaxi Yang, Jing Zhou, Jingren Zhou, Junyang Lin, Kai Dang, Keqin Bao, Kexin Yang, Le~Yu, Lianghao Deng, Mei Li, Mingfeng Xue, Mingze Li, Pei Zhang, Peng Wang, Qin Zhu, Rui Men, Ruize Gao, Shixuan Liu, Shuang Luo, Tianhao Li, Tianyi Tang, Wenbiao Yin, Xingzhang Ren, Xinyu Wang, Xinyu Zhang, Xuancheng Ren, Yang Fan, Yang Su, Yichang Zhang, Yinger Zhang, Yu~Wan, Yuqiong Liu, Zekun Wang, Zeyu Cui, Zhenru Zhang, Zhipeng Zhou, and Zihan Qiu.
\newblock {Qwen3 Technical Report}, 2025.
\newblock URL \url{https://arxiv.org/abs/2505.09388}.

\bibitem[Yang et~al.(2024)Yang, Li, Misu, Yao, Cui, Gong, Hawblitzel, Lahiri, Lorch, Lu, Yang, Zhou, and Lu]{AutoVerus}
Chenyuan Yang, Xuheng Li, Md~Rakib~Hossain Misu, Jianan Yao, Weidong Cui, Yeyun Gong, Chris Hawblitzel, Shuvendu~K. Lahiri, Jacob~R. Lorch, Shuai Lu, Fan Yang, Ziqiao Zhou, and Shan Lu.
\newblock Autoverus: Automated proof generation for rust code.
\newblock \emph{CoRR}, abs/2409.13082, 2024.
\newblock URL \url{https://doi.org/10.48550/ARXIV.2409.13082}.

\bibitem[Yang \& Deng(2019)Yang and Deng]{CoqGym}
Kaiyu Yang and Jia Deng.
\newblock {Learning to Prove Theorems via Interacting with Proof Assistants}.
\newblock In Kamalika Chaudhuri and Ruslan Salakhutdinov (eds.), \emph{Proceedings of the 36th International Conference on Machine Learning}, volume~97 of \emph{Proceedings of Machine Learning Research}, pp.\  6984--6994. PMLR, 09--15 Jun 2019.
\newblock URL \url{https://proceedings.mlr.press/v97/yang19a.html}.

\bibitem[Yang et~al.(2023)Yang, Swope, Gu, Chalamala, Song, Yu, Godil, Prenger, and Anandkumar]{yang2023leandojo}
Kaiyu Yang, Aidan Swope, Alex Gu, Rahul Chalamala, Peiyang Song, Shixing Yu, Saad Godil, Ryan~J Prenger, and Animashree Anandkumar.
\newblock {LeanDojo: Theorem Proving with Retrieval-Augmented Language Models}.
\newblock In A.~Oh, T.~Naumann, A.~Globerson, K.~Saenko, M.~Hardt, and S.~Levine (eds.), \emph{Advances in Neural Information Processing Systems}, volume~36, pp.\  21573--21612. Curran Associates, Inc., 2023.
\newblock URL \url{https://proceedings.neurips.cc/paper_files/paper/2023/file/4441469427094f8873d0fecb0c4e1cee-Paper-Datasets_and_Benchmarks.pdf}.

\bibitem[Z3(2025)]{Z3GuideArithmetic}
Z3.
\newblock Arithmetic --- online {Z3} guide, 2025.
\newblock URL \url{https://microsoft.github.io/z3guide/docs/theories/Arithmetic}.
\newblock Accessed: 2025-11-19.

\bibitem[Zheng et~al.(2022)Zheng, Han, and Polu]{MiniF2F}
Kunhao Zheng, Jesse~Michael Han, and Stanislas Polu.
\newblock {miniF2F: a cross-system benchmark for formal Olympiad-level mathematics}.
\newblock In \emph{The Tenth International Conference on Learning Representations, {ICLR} 2022, Virtual Event, April 25-29, 2022}. OpenReview.net, 2022.
\newblock URL \url{https://openreview.net/forum?id=9ZPegFuFTFv}.

\bibitem[Zhong et~al.(2025)Zhong, Zhu, Tian, and Si]{RAGVerus}
Sicheng Zhong, Jiading Zhu, Yifang Tian, and Xujie Si.
\newblock {RAG-Verus: Repository-Level Program Verification with LLMs using Retrieval Augmented Generation}.
\newblock \emph{CoRR}, abs/2502.05344, 2025.
\newblock URL \url{https://doi.org/10.48550/ARXIV.2502.05344}.

\end{thebibliography}
